\documentclass[journal]{IEEEtran}

\usepackage{graphics} 
\usepackage{epsfig} 
\usepackage{mathptmx} 
\usepackage{times} 
\usepackage{amsmath} 
\usepackage{amssymb}  

\usepackage{subcaption} 
\usepackage{listings}
\usepackage{algorithm}
\usepackage{algorithmic}

\usepackage{siunitx}
\usepackage[normalem]{ulem}

\usepackage{xcolor}
\usepackage{soul}

\usepackage{hyperref}

\usepackage[
backend=biber,  
style=ieee,
sorting=none
]{biblatex}
\title{Avoiding Dense and Dynamic Obstacles in Enclosed Spaces: Application to Moving in Crowds}

\author{Lukas Huber$^{1}$ $\cdot$ Jean-Jacques Slotine $^2$ $\cdot$ Aude Billard $^1$ \vspace{-2.3ex}
\thanks{*This work was funded in part by the EU ERC grant SAHR.}
\thanks{$^{1}$ LASA Laboratory, Swiss Federal School of Technology in Lausanne - EPFL, Switzerland. \tt $\{$lukas.huber;aude.billard$\}$@epfl.ch }
\thanks{$^{2}$ Nonlinear Systems Laboratory,  Massachusetts Institute of Technology, USA. \tt jjs@mit.edu}   
}

\begin{document}
\newcommand{\vect}[1]{\mathbf{#1}}
\newcommand{\matr}[1]{\mathbf{#1}}

\newcommand{\editcolor}[1]{\color{black}{#1} \color{black}}

\maketitle
\begin{abstract}
This paper presents a closed-form approach to constraining a flow within a given volume and around objects. The flow is guaranteed to converge and to stop at a single fixed point. \\
The obstacle avoidance problem is inverted to enforce that the flow remains enclosed within a volume defined by a polygonal surface. We formally guarantee that such a flow will never contact the boundaries of the enclosing volume or obstacles. It asymptotically converges towards an attractor. We further create smooth motion fields around obstacles with edges (e.g., tables). Both obstacles and enclosures may be time-varying, i.e., moving, expanding, and shrinking. \\
The technique enables a robot to navigate within enclosed corridors while avoiding static and moving obstacles. It was applied on an autonomous robot (QOLO) in a  static complex indoor environment and tested in simulations with dense crowds.
The final proof of concept was performed in an outdoor environment in Lausanne. The QOLO-robot successfully traversed a marketplace in the center of town in the presence of a diverse crowd with a non-uniform motion pattern.
\end{abstract}

\section{Introduction} \label{sec:introduction}
Robots navigating in human-inhabited environments will encounter disturbances constantly, for instance, when autonomous delivery robots drive around pedestrians. The robot must have a flexible control scheme to avoid collisions. As the number of obstacles increases and their motion becomes less predictable, the robot needs to reevaluate its path within milliseconds to prevent a crash while moving actively towards its goal. \\
Control using dynamical systems (DS) is ideal for addressing such situations. In contrast to classical path planning, the control law is closed-form, hence requires no replanning, and can ensure impenetrability of obstacles \cite{feder1997real, khansari2011learning}. DS offer stability and convergence guarantees in addition to the desired on-the-fly reactivity.
\subsection{Related Work}
\subsubsection{Sampling Based Exploration}
Sampling algorithms such as probabilistic road map (PRM) \cite{kavraki1996probabilistic} or the rapidly exploring random trees (RRT) \cite{lavalle2000rapidly} can find a path in cluttered environments. They are computationally intensive, and early approaches were limited to static environments. In \cite{brock2002elastic}, online (partial) replanning and elastic-band methods deform the path locally. This allowed adapting to dynamic environments. Reinforcement learning allowed the authors in \cite{zhang2013probabilistic} to learn from previously explored PRM-paths and efficiently adapt to dynamic obstacles. However, the switching often comes with the theoretical loss of global convergence \cite{vannoy2008real}. 
Recent work uses customized circuitry on a chip for onboard, fast global sampling and evaluation \cite{murray2016robot}. This method allows fast replanning, but a customized chip design is required for each robot configuration.

\subsubsection{Real-Time Optimization}
With improvements in hardware and computational speed, optimization algorithms such as model predictive control (MPC) have become feasible for onboard use in dynamic path planning, and obstacle avoidance \cite{ji2017path}. MPC has been used to control non-holonomic robots in environments with multiple convex obstacles \cite{zhang2020optimization}. Most optimization methods can not guarantee convergence to a feasible solution at runtime. \\
Power diagrams were used to identify the robot's collision-free, convex neighborhood, and an associated, well-known convex optimization problem generates a continuous flow \cite{arslan2016exact}. This method is limited to convergence for convex obstacles with almost spherical curvature. \\
Control barrier functions (CBFs) and Control Lyapunov Functions (CLFs) were united through the use of quadratic programming (QP) to create collision-free paths in \cite{ames2014control}. The convergence constraint was softened to ensure the feasibility of the optimization problem. As a result, full convergence cannot be guaranteed anymore. In \cite{reis2020control}, local minima were overcome by introducing a (virtual) orientation state. This orientation state introduces a dependence on the history of the QP problem. Additionally, the authors do not address the challenge of finding an appropriate Lyapunov-candidate. \\
In \cite{notomista2021safety} a diffeomorphic transformation to a sphere-world is used, which is based on \cite{rimon1991construction}. The method introduces a dependence on its history through virtual obstacle positions and radii. Furthermore, the construction of the diffeomorphism requires full knowledge about the space.
 
\subsubsection{Learning Based Approaches}
With the rising popularity of machine learning in the past years, these algorithms have been applied to sensor data to infer data-driven control \cite{michels2005high} but this method cannot ensure impenetrability. Other approaches use neural networks on a circular representation of crowds to create steering laws but cannot guarantee convergence \cite{long2017deep}.

\subsubsection{Velocity Obstacles}
Velocity obstacles extend the obstacle shape by its potential future positions based on the current velocity \cite{fiorini1998motion}. Velocity obstacles allow safe navigation in dynamic environments. They were successfully applied in multi-agent scenarios \cite{van2008reciprocal} and extended to include acceleration, and non-holonomic constraints of the agent \cite{wilkie2009generalized}. The velocity obstacle approach often conservatively limits the workspace.

\subsubsection{Artificial Potential Fields and Navigation Functions}\label{sec:artificial_potential_fields}
Artificial potential fields were used to create collision-free trajectories \cite{khatib1986real}, but they are prone to local minima. In \cite{koditschek1990robot} artificial potential fields for sphere-worlds were designed to have only a global minimum. A diffeomorphic transformation was introduced to map \textit{star-worlds} to \textit{sphere-worlds} \cite{rimon1991construction}, and extended to include \textit{trees-of-stars} \cite{rimon1992exact}. The tuning of critical parameters needs the knowledge of the whole space. Hence, in practice, full convergence is difficult to achieve. \\
More recent approaches introduce artificial potential fields with only the global minimum for more general environments \cite{loizou2011closed, paternain2017navigation}. Automated approximation of the tuning parameter has been proposed \cite{loizou2017navigation}, but it does not generalize easily to dynamic environment. In \cite{kumar2019navigation}, full convergence is ensured around ellipse obstacles through quadratic potential functions. Learning methods were used to tune the hyper-parameters of potential fields to obtain human-inspired behavior for obstacle avoidance \cite{duan2020learning}. \\
Danger-fields were used in \cite{lacevic2013safety} to ensure collision avoidance for robot arms through repulsion from dynamic obstacles. The approach was extended by guiding the motion through an artificial potential field in \cite{zanchettin2015passivity}. The design of the artificial potential field remains a challenge for this method.
Dynamic reference points help to reduce the probability of converging to a local minimum for potential fields \cite{zhang2019distributed}.

\subsubsection{Harmonic Potential Fields}
Harmonic potential functions are interesting as they guarantee that no topologically critical points arise in free space. In \cite{connolly1990path}, the harmonic potential functions are evaluated numerically to overcome the challenge of finding them analytically. Closed-form harmonic potential functions can be generated by approximating the obstacles through linear \textit{panels} \cite{kim1992real}.  This linear approximation applies to concave obstacles but is limited to two-dimensional environments \cite{feder1997real}. In \cite{guldner1993sliding}, known harmonic potential functions are interpolated to navigate in more complex environments. A closed-form solution to harmonic potential flow around simple obstacles was presented in \cite{khansari2012dynamical}. The work allows to avoid moving obstacles but is limited to convex. In \cite{saveriano2014distance}, the approach was extended to concave obstacles by using a discrete, sensor-based representation. Closed-form approaches using harmonic potential fields are often simplified to a circular world or require high (close to circular) curvature. \\
In \cite{huber2019avoidance}, a dynamic reference point was introduced to ensure convergence for star-shaped environments. However, the approach was not able to handle boundaries or deforming obstacles.

\subsection{Contributions}
This paper addresses the need for a reactive (closed-form) obstacle avoidance approach with formal guarantees of impenetrability to handle highly dynamic environments and realistic obstacles, such as obstacles with sharp edges. To this end, this paper extends our previous work \cite{huber2019avoidance}, in which we presented a closed-form obstacle avoidance approach guaranteed to not penetrate smooth concave, albeit star-shaped, obstacles. We present three novel theoretical contributions: \\
1) We invert the obstacle description to ensure that the robot moves within the enclosed space defined by the boundary while preserving convergences guarantees towards an attractor. This boundary may represent walls, furniture, or even joint limits of manipulators (Sec.~\ref{sec:Inverted_obstacle}). \\
2) We extend the approach to handle {\em nonsmooth} surfaces, i.e., obstacles with sharp edges. The novelty comes from creating a smooth dynamical system around an obstacle without approximation of the curvature (Sec.~\ref{sec:nonsmooth}). \\
3) We show that the approach can be extended to tackle dynamic environments, with obstacles that have {\em deforming} shapes (Sec.~\ref{sec:dynamic_environments}).

\subsubsection{Implementation and Technical Contributions}
We validate these contributions with a wheelchair robot moving in a simulated crowd of pedestrians and an office environment with real furniture (Sec.~\ref{sec:empricvalValidation}). \\
For the real-world implementation, we present three technical contributions: \\
T.1) We extend the modulation parameters of the obstacle avoidance with surface friction and repulsive value (Sec.~\ref{sec:obstacle_avoidance}). They allow an agent to move slower and further away from obstacles, respectively. The extensions result in more \textit{cautious} behavior. \\
T.2) We propose an approach on how to use the dynamical obstacle avoidance in combination with a robot arm (Sec.~\ref{sec:robot_arm}). The safe joint velocity is evaluated based on the proximity function combined with the obstacle avoidance algorithm. \\
T.3) We introduce a general directional weighting for (ensured) non-trivial summing of vector fields and corresponding gradient descent (Sec.~\ref{sec:dir_weighted_mean}).

\editcolor{\subsection{Notation}
The state variable $\xi \in \mathbb{R}^d$ defines the state of a robotic system. If not mentioned otherwise, $\xi$ will refer to the Cartesian position of the agent. \\
  The variable $\pi$ is used as the circle constant throughout the paper. \\
  Superscripts are used to denote the name of variables and subscripts for enumeration. \\
  Bold-face Latin characters describe vectors and matrices. Vectors defined by Greek characters are explicitly defined. \\
  The brackets $ \langle \cdot , \cdot \rangle $ denote the dot product of two vectors. \\
  The $\times$-operator indicates the cross product. \\
  The square brackets indicate elements of a vector, e.g., $\xi_{[1]}$ denotes the first element of $\xi$. Double dots within the brackets indicate a sub-vector up to the specified number, e.g., $\xi_{[1:2]}$ is a vector of the first two elements of $\xi$.}
 
\section{Obstacle Avoidance Formulation}  \label{sec:obstacle_avoidance}
Dynamical system based obstacle avoidance has been proposed by the authors in \cite{huber2019avoidance}. We restate previous definitions, and introduce two new extensions: a \textit{friction parameter} (Sec.~\ref{sec:surface_friction}) and a \textit{repulsion parameter} (Sec.~\ref{sec:repulsive_eigenvalue}).

\subsection{Dynamical Systems}
This work focuses on motion towards a goal of an autonomous dynamical system, i.e. $\lim_{t-\rightarrow \infty} f(\xi_a) = 0$. The most direct dynamics towards the attractor is a linear dynamical system of the form:
\begin{equation}
  \vect f(\xi) = - k (\xi - \xi^a)
\end{equation}
where $k \in \mathbb{R}$ is a scaling parameter. 
The attractor $\xi^a$ is visualized throughout this work as a star: $*$ and with set  $k=1$.

\subsection{Obstacle Description}
Each obstacle has a continuous distance function $\Gamma(\xi): \mathbb{R}^d  \mapsto \mathbb{R}_{\geq 0}$, which allows to distinguish three regions:
\begin{align}
  &\text{Free points:}&  \qquad & \mathcal{X}^f = \{\xi \in \mathbb{R}^d: \Gamma(\xi) > 1 \} \nonumber \\ 
  &\text{Boundary points:}&  \qquad & \mathcal{X}^b = \{\xi \in \mathbb{R}^d: \Gamma(\xi) = 1 \} \label{eq:levelFunc} \\
  &\text{Interior points:}&  \qquad & \mathcal{X}^o = \{ \xi \in \mathbb{R}^d \setminus (  \mathcal{X}^f \cup \mathcal{X}^b ) \} \nonumber
\end{align}

\subsubsection{Reference Point}
For each obstacle $i$ a reference point is chosen such that it lies within the kernel of the obstacle: $\xi^r_i \in \mathcal{X}^o_i$.\footnote{The kernel of a star-shaped obstacle defines the region from which any surface point is visible \cite{lee1979optimal}} The reference direction is defined as:
\begin{equation}
{\vect r_i}(\xi) = {\left( \xi - \xi^r_i \right)}/{\|\xi - \xi^r_i \|}   
\quad \forall \xi \in \mathbb{R}^d \setminus \xi^r_i 
\label{eq:reference_direction}
\end{equation}
The reference point is visualized throughout this work as a cross $+$.

\subsubsection{Distance Function}
By construction, the distance function $\Gamma(\cdot)$ increases monotonically in radial direction and has a continuous first-order partial derivative ($C^1$ smoothness). Here, we define the general distance function as:
\begin{equation}
  \Gamma^o( \xi) = \left( \| \xi-\xi^r\| / R(\xi)\right)^{2p} \quad \forall \xi \in \mathbb{R}^d \setminus \xi^r   \label{eq:distFunction_example}
\end{equation}
with the power coefficient $p \in \mathbb{N}_+$. The local radius $R(\xi) = \| \xi^b - \xi^r \|$ is a function of the local boundary point, which is defined as:
\begin{equation}
  \xi^b = b \vect r (\xi) + \xi^r  \quad \text{such that} \quad b>0 \, , \;\; \xi^b \in \mathcal{X}^{b} \label{eq:boundary_point}
\end{equation}

\subsection{Obstacle Avoidance through Modulation}
Real-time obstacle avoidance is obtained by applying a dynamic modulation matrix to a dynamical system $\vect f(\xi)$:
\begin{equation}
  \dot{\xi} = \matr{M}(\xi) \vect f(\xi)
  \qquad \text{with} \quad
  \matr{M} ( \xi) = \matr{E}(\xi) \matr D(\xi) \matr{E}(\xi)^{-1} \label{eq:modDS_app}
\end{equation}
The modulation matrix is composed of the basis matrix:
\begin{equation}
  \matr {E} (\xi) =
  \left[ {\vect r }(\xi) \;\; \vect e_1(\xi) \;\; .. \;\; \vect{e}_{d-1}(\xi) \right]
   \label{eq:basisMatr}
\end{equation}
which has the orthonormal tangent vectors $\vect e_i(\xi)$ evaluated at the boundary point $\xi^b$ given in (\ref{eq:boundary_point}). \\
The diagonal eigenvalue matrix is given as:
\begin{equation}
  \matr D(\xi) =
  \textbf{diag}
  \left(
    \lambda^r(\xi) ,
    \lambda^e(\xi) , \,
     \hdots ,
     \lambda^e( \xi)
     \right)
  \label{eq:eigVecMatr}
\end{equation}
We set the eigenvalues to
\begin{equation}
  \lambda^r(\xi) = 1 - {1}/{\Gamma(\xi)}^{1/\rho} \qquad \lambda^e(\xi) = 1 + {1}/{\Gamma(\xi)}^{1/\rho}
  \label{eig:valuesFluid}
\end{equation}
with the reactivity factor $\rho \in \mathbb{R}_{>0}$ and the distance function $\Gamma(\xi)$ from (\ref{eq:distFunction_example}). In this work, we simply choose $\rho = 1$.

\subsection{Multiple Obstacles}
In the presence of multiple obstacles, the velocity is modulated for each obstacle individually as described in (\ref{eq:modDS_app}). The final velocity is obtained by taking the weighted directional mean of the individual velocities, see Sec.~\ref{sec:dir_weighted_mean}.

\subsection{Surface Friction Imitation} \label{sec:surface_friction}
\begin{figure}[t]
\centering
\includegraphics[width=1.0\columnwidth]{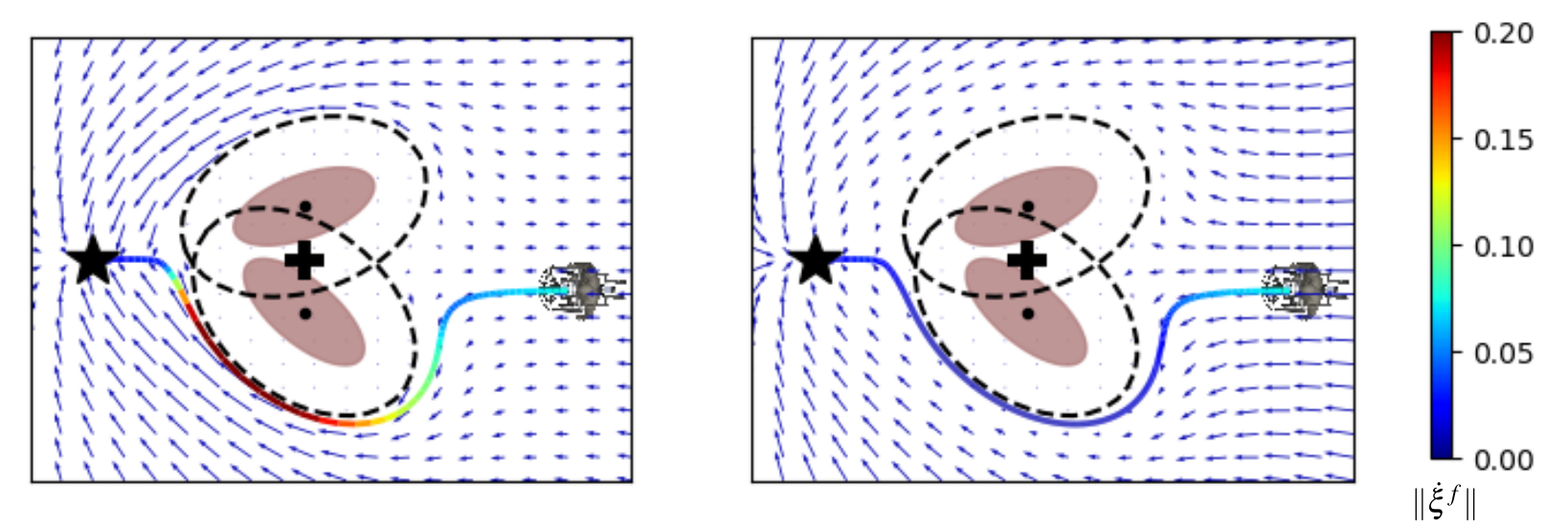}
\caption{The velocity obtained with the isometric inspired eigenvalue (left) results in accelerations close to the obstacle. The modulation inspired by surface friction (right) reduces the velocity with decreasing distance to the obstacle. The black star is the attractor $\xi^a$, and the black cross is the (shared) reference point $\xi^r$.}
\label{fig:eigenvalues_surface_friction}
\end{figure}

The choice of eigenvalues in (\ref{eig:valuesFluid}) is inspired by the harmonic potential flow, i.e., the description of the potential flow of an incompressible fluid. The incompressibility constraint forces the velocity to increase in regions where the flow is pushed around the obstacle, i.e., the eigenvalues in the tangent direction increase. This leads to acceleration close to the surface (see Fig.~\ref{fig:eigenvalues_surface_friction}). \\
We propose to mimic surface friction, i.e., slowing down in tangent direction close to an obstacle ($\lim_{\Gamma \rightarrow 1} \xi^e = 0$). 
A friction parameter $\lambda^f(\xi)$ ensures the slowing down close to the surface. The friction dynamics $\dot{\xi}^f$ are obtained by applying the factor to tangent and reference direction as follows:
\begin{equation}
  \dot{\xi}^f = \lambda^f(\xi) \frac{\| \vect f(\xi) \|}{\| \dot{\xi} \|} \dot{\xi}
  \qquad \text{with} \quad
  \lambda^f(\xi) = 1 - 1/\Gamma(\xi)
\end{equation}
where the $\dot{\xi}$ is the modulated velocity from (\ref{eq:modDS_app}).

\subsection{Repulsive Eigenvalue} \label{sec:repulsive_eigenvalue}
\begin{figure}[t]\centering
\centering
\includegraphics[width=0.9\columnwidth]{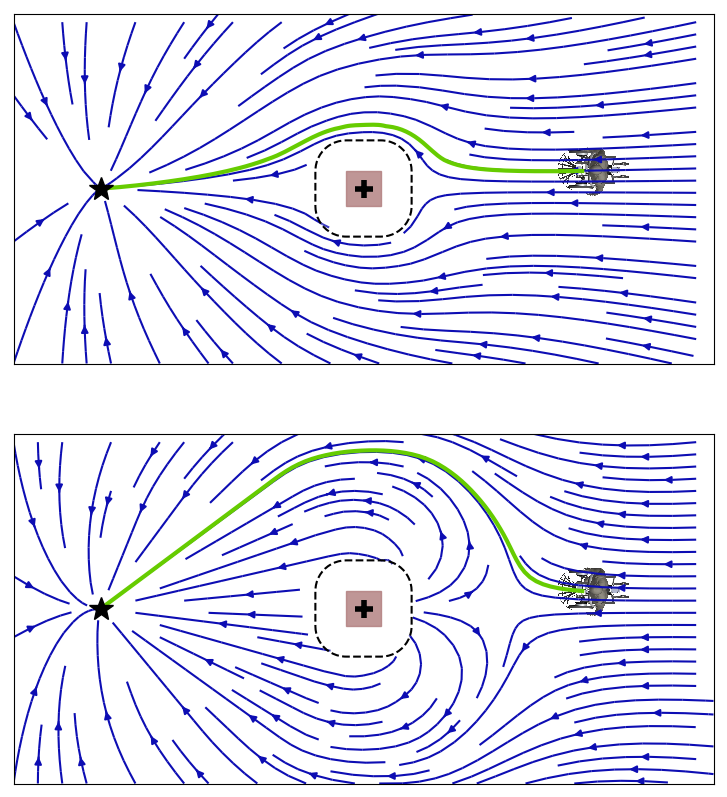}
\caption{A repulsion coefficient of $c^{\mathrm{rep}}=1$ results in strictly positive eigenvalues (top). An increased distance to the obstacle is obtained with higher repulsion coefficients, such as  $c^{\mathrm{rep}}=2$ (bottom). }
\label{fig:eigenvalues_repulsive_comparison}
\end{figure}

The eigenvalues in reference direction given in (\ref{eig:valuesFluid}) are always positive, as follows from $\Gamma(\xi) \geq 1$. This results in eigenvalues in the range of $\lambda^r(\xi) \in [0, 1]$, hence reduces the velocity in radial direction. \\
Conversely, active repulsion can be achieved through negative eigenvalues, i.e., $\lambda^r(\xi) < 0$. Active repulsion increases the distance by which the an agents avoids the obstacle (Fig.~\ref{fig:eigenvalues_repulsive_comparison}). Since the repulsive eigenvalues are incorporated in the modulation matrix in (\ref{eq:modDS_app}), it is ensured that attractors are preserved. \\
The eigenvalue in radial direction is defined for repulsive obstacles as:
\begin{equation}
\lambda^r(\xi) =
\begin{cases}
1 - \left({c^{\mathrm{rep}}}/{\Gamma(\xi)}\right)^{1/\rho}  & \text{if} \; \langle \vect f(\xi), \vect r \rangle < 0  \\
1 & \text{otherwise}
\end{cases}
\end{equation}
with the repulsive coefficient $c^{\mathrm{rep}} \geq 1$. A repulsive coefficient $c^{\mathrm{rep}}=1$ corresponds to no repulsion. Note, that the repulsive eigenvalues are coupled with no \textit{tail-effect}, i.e. $\lambda^r(\xi) = 0$ in the wake of an obstacle (see \cite{khansari2012dynamical}).

\section{Inverted Obstacle Avoidance} \label{sec:Inverted_obstacle}
An autonomous robot often encounters scenarios where it has boundaries that it cannot pass. This might be a wall for a wheeled robot or the joint limits for a robot arm. These constraints can be interpreted as staying within an obstacle, where the obstacle represents the limits of the free space.

\subsection{Distance Inversion} \label{sec:distance_inversion}
The distance function $\Gamma(\xi)$ from (\ref{eq:distFunction_example}) can be evaluated within the obstacle $\xi \in \mathcal{X}^o$.\footnote{In the classic obstacle avoidance case, this is of no use, since theoretically the obstacle does never reach the boundary \cite{huber2019avoidance}, and practically an \textit{emergency} control has to be applied in this case.}
For interior points, our boundary function is monotonically decreasing along the radial direction, i.e. the Lie derivative with respect to the reference direction is positive. Furthermore it is bounded. This can be written as:
\begin{equation*}
  {L}_{\vect r} \Gamma = 
  \langle \frac{\partial \Gamma(\xi)}{\partial \vect r(\xi)}, \vect r(\xi) \rangle > 0,  \hspace{0.5cm} \;\; \Gamma(\xi) \in \left[0,1\right[ \quad \forall \; \mathcal{X}^o \setminus \xi^r
\end{equation*}

\begin{figure}[t]
\centering
\includegraphics[width=0.9\columnwidth]{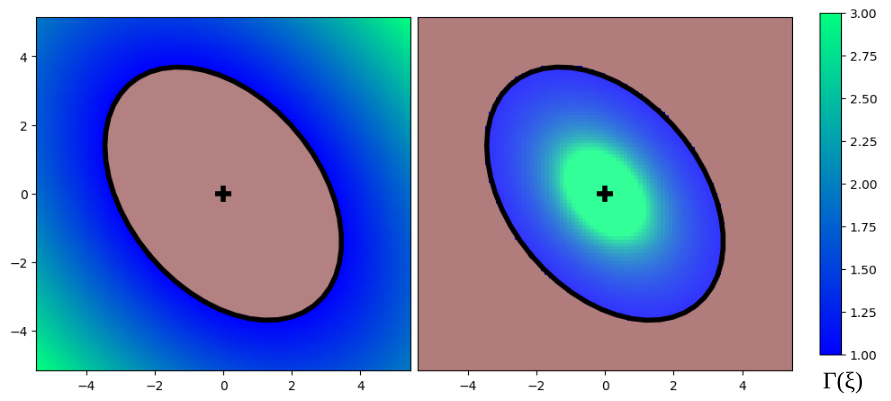}
\caption{Forward and inverted $\Gamma$-function for the same shape. The brown region marks the inside of the obstacle and wall, respectively.}
\label{fig:gamma_inversion}
\end{figure}

We consider the obstacle boundary $\mathcal{X}^b$ as the description of an enclosing hull. It follows that the interior points of the classical obstacle become points of free space of the enclosing hull and vice versa. Boundary points stay boundary points. We define the distance function of wall-obstacles as the inverse of the obstacle distance function:
\begin{equation}
  \Gamma^w(\xi) = 1/\Gamma^o = \left( R(\xi)/ \|\xi-\xi^r \| \right)^{2p} \qquad \forall \; \mathbb{R}^d \setminus \xi^r \label{eq:inverse_gamma}%
\end{equation}
This new distance function fulfills the condition for the three regions as given in (\ref{eq:levelFunc}). The distance function $\Gamma^w$ is now monotonically decreasing along radial direction and reaches infinity at the reference point, i.e., $\lim_{\xi \rightarrow \xi^r}\Gamma^w(\xi) \rightarrow \infty$ (see Fig.~\ref{fig:gamma_inversion}).

\subsection{Modulation Matrix}
The modulation matrix, defined in (\ref{eq:modDS_app}), consists of the eigenvalue $\matr D(\xi)$ and basis matrix $\matr E(\xi)$. For inverted obstacles, the diagonal eigenvalue matrix from (\ref{eq:eigVecMatr}) is a function of the inverted distance function $\Gamma^w(\xi)$. \\
Conversely, the basis matrix is constant along the radial direction. It can be evaluated everywhere (including the interior of a boundary) except at the reference point ($\xi = \xi^r$). Since the reference direction from (\ref{eq:reference_direction}) is a zero vector, no orthogonal basis is defined. However, the distance value $\Gamma^w(\xi^r)$ reaches infinity, hence from (\ref{eq:eigVecMatr}) we get that the diagonal matrix is equal to the identity matrix:
\begin{equation}
  \Gamma^w(\xi^r) \rightarrow \infty
  \quad \Rightarrow \quad
  \matr D(\xi^r) = \matr I
\end{equation}
Using (\ref{eq:modDS_app}),  it follows for the modulated dynamics at the reference point:
\begin{multline}
  \matr M(\xi^r)
  = \matr E(\xi^r) \matr D(\xi^r) \matr E(\xi^r)^{-1}
  = \matr E(\xi^r) \matr I \matr E(\xi^r)^{-1}
  = \matr I \\
  \Rightarrow
  \dot \xi^r = \vect f(\xi^r)
  \label{eq:modulation_reference}
\end{multline}
The influence of modulation approaches zero when reaching the reference point. Hence the dynamical modulation is continuously defined.  \\
\\
\noindent\textbf{Theorem 1}
\textit{Consider a star-shaped enclosing wall in $\mathbb{R}^d$ with respect to a reference point inside the obstacle $\xi^r$, and a boundary $\Gamma^w(\xi)=1$ as in (\ref{eq:inverse_gamma}). Any trajectory $\{\xi\}_t$, that starts within the free space of an enclosing wall, i.e., $ \Gamma(\{\xi \}_0) > 1$ and evolves on a smooth path according to (\ref{eq:modDS_app}), will never reach the wall, i.e., $\Gamma(\{ \xi\}_t) > 1, \; t = 0..\infty$ and converges towards an attractor $\xi^a \in \mathcal{X}^f$, i.e., $\lim_{t \rightarrow \infty}\xi \rightarrow \xi^a$.}\hfill \textbf{Proof:}~see~Appendix~\ref{sec:proofTheorem1}.
\begin{figure}[t]\centering
\begin{subfigure}{.49\columnwidth} %
\centering
\includegraphics[width=\textwidth]{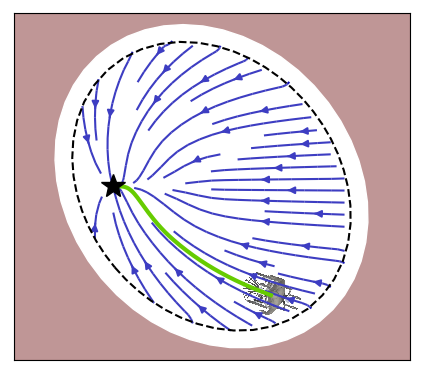}
\caption{Ellipsoid boundary}
\label{fig:ellipsoid_boundary}
\end{subfigure}%
\begin{subfigure}{.49\columnwidth} %
\centering
\includegraphics[width=\textwidth]{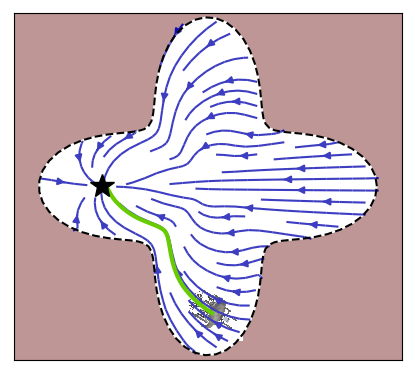}
\caption{Star-shaped boundary}
\label{fig:starshaped_boundary}
\end{subfigure}%
\caption{A smooth flow with full convergence towards the attractor (black star) can be observed within any star-shaped wall with reference point $\xi^r$ (black plus).}
\label{fig:boundary_analysis}
\end{figure}

\subsection{Guiding Reference Point to Pass Wall Gaps}
\begin{figure}[t]\centering
\centering
\includegraphics[width=1.0\columnwidth]{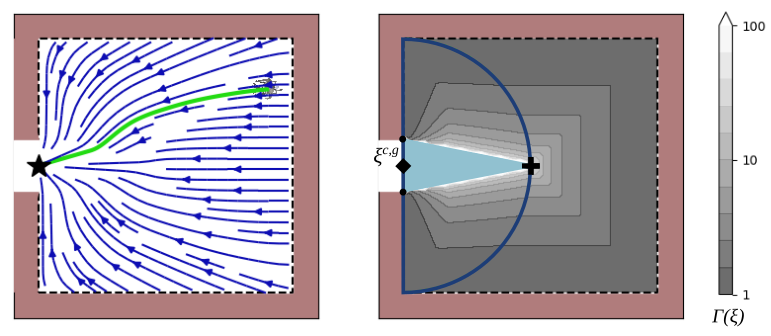}
\caption{The dynamical system (left) is not modulated in front of the gap since the $\Gamma$-function reaches infinity (right). The \textit{gap region} $\mathcal{X}^g$ is the blue region. The influence of the gap is limited by the blue half-circle around the center of the gap $\xi^{c,g}$ (black square).}
\label{fig:boundary_with_gap_subplot}
\end{figure}

In many practical scenarios, a hull entails gaps or holes through which the agent enters or exits the space (e.g., door in a room). The modulation-based avoidance slows the agent down when approaching the boundary and does not let it pass through such an exit. We introduce a \textit{guiding reference point} $\xi^g$ for boundary obstacles to counter this effect. We assume convex walls, and the robot size being smaller than the gap width. It is assumed that gap is a priori known from a map.\\
In (\ref{eq:modulation_reference}), it was shown that at the center of an inverted obstacle, the influence of the modulation vanishes. \\
Let us define the \textit{gap region} $\mathcal{X}^g$ enclosed by the lines connecting the gap edges and the reference point $\xi^r$ (see Fig.~\ref{fig:boundary_with_gap_subplot}). The guiding reference point is designed in the following manner: close to the gap, the guiding reference point is equal to the position of the evaluation, hence no influence of the modulation. Far away from the gap, the guiding reference point is equal to the reference point; hence the wall has no influence. In between the two regions, the guiding reference point is projected onto the gap region $\mathcal{X}^g$. This can be written as:
\begin{equation*}
\xi^{g} =
\begin{cases}
  \xi \quad &\text{if} \;\;\; \xi \in \mathcal{X}^g \\
  \text{argmin}_{\hat \xi \in \mathcal{X}^g} \| \xi - \hat \xi \| &\text{else if} \; \frac{\| \xi^{c,g} - \xi \|}{\| \xi^{c,g} - \xi^r \|} > 1  \\
  \xi^r \quad & \text{otherwise}
\end{cases}
\end{equation*}
with $\xi^{c,g}$ the center point of the gap (see Fig.~\ref{fig:boundary_with_gap_subplot}).

\section{Nonsmooth Surfaces} \label{sec:nonsmooth}
Human-designed environments often contain obstacles and enclosing walls with nonsmooth surfaces, e.g., a table with edges or a building with corners. An approximation with a high gradient of these surfaces can lead to undesired results. On the one hand, a smoothing of the edges increases the risk of colliding with them. On the other hand, an increased, smooth hull conservatively increases the boundary region, and certain parts of the space are not reachable with such a controller.  \\
Moreover, the obstacle avoidance algorithm applied to a surface with a high gradient can lead to a fast change of the flow. Even though the trajectories are smooth, the curvature of the flow can be high and induce fast accelerations. This can result in dangerous behavior in the presence of humans or simply exceed the robot's torque limits. 
We propose an approach to avoid obstacles with nonsmooth surfaces without smoothing the boundary. \\
A polygonal obstacle consists of $i = 1 ... N^s$ individually smooth surface planes which form a star shape in $d-1$ such that:
\begin{align}
  \mathcal{X}^s_i = \{ &\xi, \hat{\xi} \in \mathcal{X}^b,  \; \exists \, \vect n_i \, : \,
  \langle \vect n_i, \, (\xi - \hat{\xi}) \rangle = 0 \} \label{eq:non_smooth_planes}
\end{align}

\begin{figure}[t]\centering
\centering
\includegraphics[width=1.0\columnwidth]{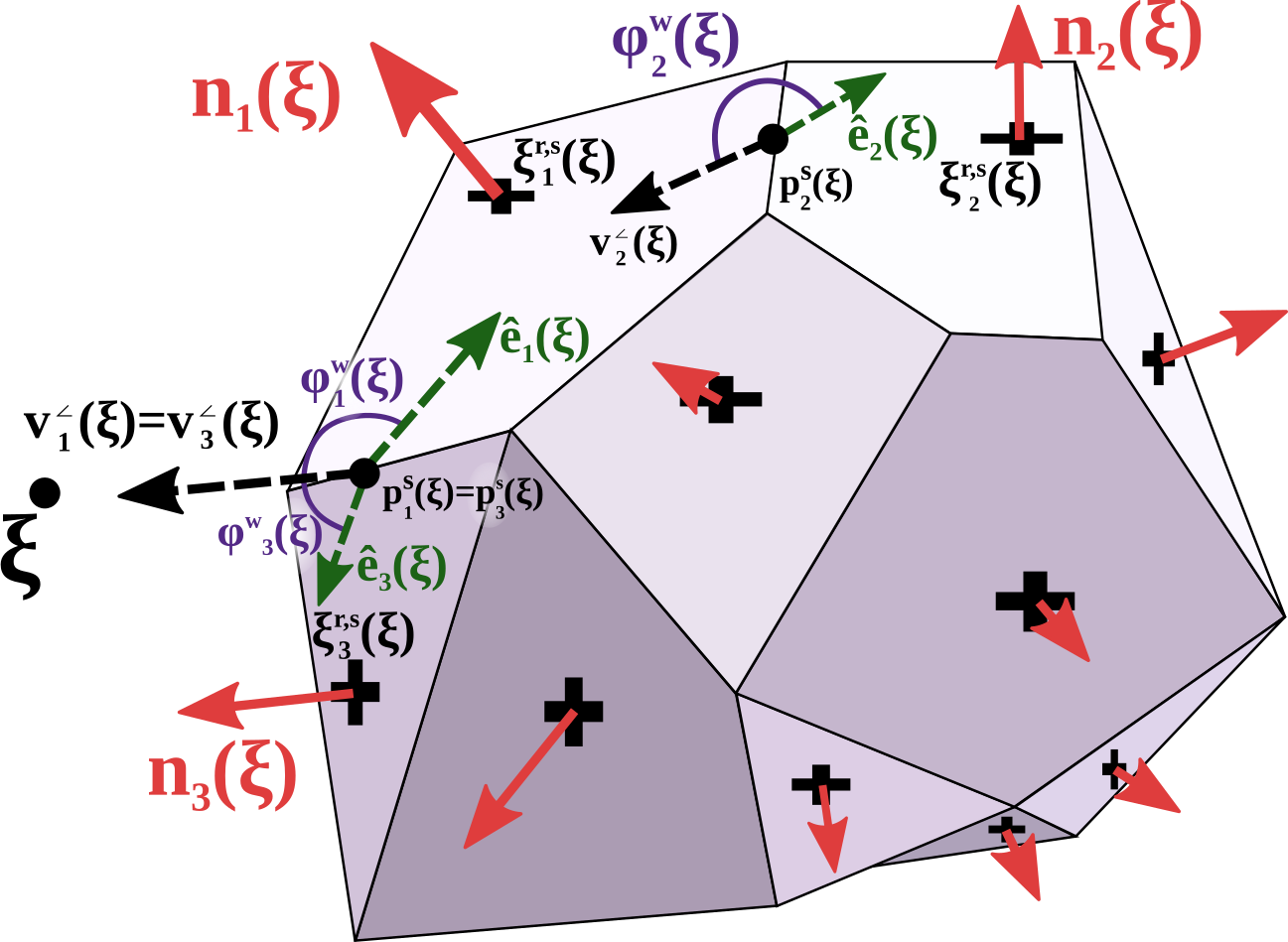}
\caption{The variables for the evaluation of the pseudo normal $\hat{\vect n}(\xi)$ of a nonsmooth star-shaped obstacle are displayed for three surface tiles. Only the surface reference point (cross) and the surface normal (red arrow) are visualized for the other tiles. The angle $\varphi_{}^w$ is evaluated for each surface at the edge-point closest to $\xi$.}
\label{fig:nonsmooth_obstacle}
\end{figure}

\subsection{Pseudo Normal Vector} \label{sec:pseudo_normal}
The normal to the surface of the obstacle is not defined continuously. As a result, the modulated flow would not be smooth. \\
We create a smoothly defined pseudo normal $\hat{\vect n}(\xi)$. It is equal to the normal on the surface of the obstacle. While far away from the obstacle, the pseudo normal approaches the reference direction, i.e.:
\begin{equation}
    \hat{\vect n}(\xi) = \vect n_i(\xi) \;\; \forall \xi \in \mathcal{X}^s_i
    \quad \text{and} \quad
    \lim_{\|\xi - \xi^r\| \rightarrow \infty} \hat{\vect n}(\xi) = \vect r(\xi)
\end{equation}

\begin{figure}[t]\centering
\begin{subfigure}{.49\columnwidth} %
\centering
\includegraphics[width=\textwidth]{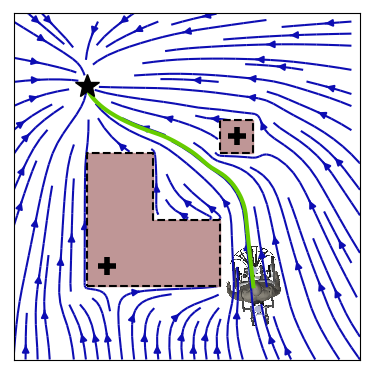}
\caption{Nonsmooth obstacles}
\label{fig:edge_obstacles_several}
\end{subfigure}%
\begin{subfigure}{.49\columnwidth} %
\centering
\includegraphics[width=\textwidth]{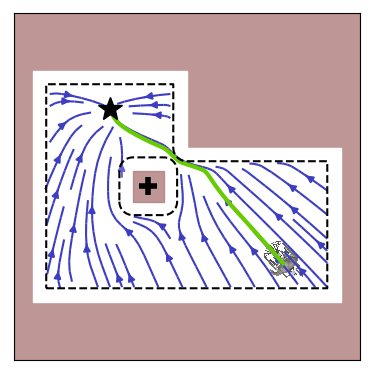}
\caption{Nonsmooth boundaries}
\label{fig:sharp_boundary_with_obstacle}
\end{subfigure}%
\caption{Nonsmooth inverted obstacles representing rooms or boundary conditions.}
\label{fig:nonsmooth_obstacles}
\end{figure}

The pseudo normal is the weighted sum of the normals, with the weights being evaluated as follows.
At first the vector from the closest edge point $\vect p_i^s$ (Fig.~\ref{fig:nonsmooth_obstacle}) to the agent's state is created:
\begin{equation}
  \vect v_i^\angle(\xi) = \xi - \vect p_i^s
  \;\;\; \text{with} \;\;\;
  \vect p_i^s = \underset{\hat \xi \in \mathcal{X}^e_i}{\mathrm{argmin}} \| \xi -  \hat \xi \| \label{eq:shortest_distance}
\end{equation}
with $\mathcal{X}^e_i$ the set of all points at the edge of a surface tile $i$.
This vector is further projected onto the surface plane:
\begin{equation}
  \hat{\vect e}_i(\xi) = \left( \vect v^\angle_i(\xi) - \langle \vect n_i (\xi), \, \vect v_i^\angle(\xi)\rangle \,  \vect n_i  (\xi) \right) \mathrm{sign} \langle \vect v_i^\angle(\xi), \, \xi - \xi^{r,s}_i  \rangle
  \nonumber
\end{equation}
The angle to the plane (Fig.~\ref{fig:nonsmooth_obstacle}) in the range $[0, \pi]$ is evaluated as:
\begin{equation}
    \varphi^{w}_i(\xi) = \arccos \left( \frac{\langle \hat{\vect e}_i(\xi), \, \vect v_i^\angle(\xi)  \rangle}{\|\hat{\vect e}_i(\xi)\|  \, \| \vect v_i^\angle(\xi)\|}\right)   \text{sign} \langle \vect n_i(\xi), \, \vect{v}_i^\angle(\xi)  \rangle
\end{equation}  
The edge-weight is evaluated as:
\begin{equation}
  \tilde w^{s}_i(\xi) =
 \begin{cases}
   \left( \frac{\pi}{\varphi^{w}_i(\xi)} \right)^{p}-1 & \text{if} \;\; \varphi^w_i(\xi) \in \, ]0, \, \pi ] \\
    0 & \text{otherwise}
  \end{cases}
  \label{eq:angle_weight}
\end{equation}
with the the weight power $p \in \mathbb{R}$, a free parameter. A lower weight power $p$ results in an increased importance for the closest polygon face compared to the other faces, we simply choose $p=3$.
The final step is the normalization of the weights:
\begin{equation}
w^s_i(\xi) =
  \begin{cases}
    \tilde w^s_i(\xi) / \sum_j \tilde w^s_j(\xi)  & \text{if} \;\; \xi \in \mathbb{R}^d \setminus \mathcal{X}^s_i \\
    1 & \text{otherwise}
  \end{cases}
  \label{eq:angle_weight_normalize}
\end{equation}
The pseudo normal is evaluated as the directional weighted mean (see Sec.~\ref{sec:dir_weighted_mean}) of normal vectors of the surface tiles $\vect n_i(\xi)$, the weights $w^s_i$, and with respect to the reference direction $\vect r(\xi)$. \\
The basis matrix from (\ref{eq:basisMatr}) is redefined for nonsmooth surfaces as:
\begin{equation}
    \matr {E} (\xi) =
  \left[ {\vect r }(\xi) \;\; \hat{\vect e}_1(\xi) \;\; .. \;\; \hat{\vect{e}}_{d-1}(\xi) \right] \label{eq:basis_matrix_nonsmooth}
\end{equation}
with $\hat{\vect e}_i(\xi)$ the orthonormal basis to $\hat{\vect n}(\xi)$. The resulting smooth vectorfield can be observed in Fig.~\ref{fig:nonsmooth_obstacles}.

\subsection{Inverted Obstacles} \label{sec:inverted_polygon}
For an inverted obstacle (Sec.~\ref{sec:Inverted_obstacle}) the pseudo normal is evaluated at the mirrored position $\xi^{\mathrm{mir}}$. It is obtained by flipping the current robot state $\xi$  along the reference direction $\vect r(\xi) = \xi-\xi^r$ onto the other side of the boundary (\ref{eq:distFunction_example}):
\begin{equation}
\xi^{\mathrm{mir}} =  \Gamma(\xi)^2  \left( \xi-\xi^r \right) + \xi^r
\end{equation}
The mirrored position allows the evaluation of the distance function as described in Sec.~\ref{sec:pseudo_normal}. Further, the inverted obstacle is treated as described in Sec.~\ref{sec:Inverted_obstacle}. This allows avoiding nonsmooth obstacles and boundaries in Fig.~\ref{fig:boundary_analysis}.
\\
\\ \noindent\textbf{Theorem 2}
\textit{Consider a polygon composed of $N^s$ surfaces as given in $(\ref{eq:non_smooth_planes})$ or alternatively an inverted polygon as described Sec.~\ref{sec:inverted_polygon}. Any trajectory $\{\xi\}_t$, that starts in free space, i.e. $ \Gamma(\{\xi \}_0) > 1$ and evolves on a smooth path according to (\ref{eq:modDS_app}), will never reach the surface, i.e. $\Gamma(\{ \xi\}_t) > 1, t = 0..\infty$ and will converge towards the attractor as long as it is placed outside all obstacles, i.e. $\lim_{t \rightarrow \infty}\xi \rightarrow \xi^a \in \mathcal{X}^f$. }   \hfill \textbf{Proof:}~see~Appendix~\ref{sec:proofTheorem2}.\\


\subsection{Implementation}
Pseudo-normals are useful for surroundings with sharp boundaries, for example, pieces of furniture with edges or maps of buildings with sharp corners. These cases consist of polygons with a small number of faces. They allow the evaluation of the algorithm in real-time. It is designed for scenarios with obstacles known from previously learned libraries or a known map retrieved at runtime, such as the tracker of \cite{jia2020dr}. \\
Conversely, if the input data is a point cloud (e.g., Lidar) or an obstacle-mesh, the surface can be approximated using a standard regression technique. The surface normal can be obtained directly by taking the derivative or by learning the normal at regression time \cite{ao2021spinnet}.

\section{Dynamic Environments} \label{sec:dynamic_environments}
In changing environments with moving or deforming obstacles the system is modulated with respect to the relative velocity as:
\begin{equation}
  \dot{\vect \xi} = { \matr{M}(\xi)} \left( \vect f (\xi) - \dot{\tilde \xi}^{\mathrm{tot}} \right) + \dot{\tilde \xi}^{\mathrm{tot}} \label{eq:dynamic_environments}
\end{equation}
The local, relative velocity is summed up over all obstacles
\begin{equation}
  \dot{\tilde \xi}^{\mathrm{tot}} = \sum_{o=1}^{N_o} w^{{l}}_o \dot{\tilde \xi}_o
    \label{eq:relative_obstacle_velocity}
\end{equation}
with the dynamic weight being a function of the distance $\Gamma(\xi)$:
\begin{equation*}
  {w}^{{l}}_o = \frac{w^{l}_o}{\sum_o{\tilde w^{{l}}_o}}
  \quad \text{with} \;\;
  \tilde w^{{l}}_o = \frac{1}{\Gamma_o(\xi)-1}
  \quad
  \forall \; \Gamma_o(\xi) > 1
  \label{eq:obtacle_weight}
\end{equation*}
The relative velocity consists of the obstacle's velocity $\dot{\tilde{\xi}}^v_o$ and deformation $\dot{\tilde{\xi}}^d_o$:
\begin{equation}
  \dot{\tilde \xi}_o =  \dot{\tilde{\xi}}^v_o + \dot{\tilde{\xi}}^d_o
\end{equation}
Note that avoiding dynamic obstacles is not only a modulation of the DS, i.e., a matrix multiplication. This can result in the velocity at the attractor being non-zero, even though the initial dynamical system has zero value there: $\vect f(\xi^a) = \vect 0$. \\
For the rest of this section, we will assume the application to each obstacle implicitly without using the subscript $(\cdot)_{o}$.

\subsection{Moving Obstacles} \label{sec:movingObstacles}
The relative velocity of a moving obstacle is obtained similarly to \cite{khansari2012thesis}:
\begin{equation}
  \dot{\tilde{\xi}}^v = \dot{\vect \xi}^{L,v} + \dot{ \vect{\xi}}^{R,v} \times \tilde{\vect \xi} \label{eq:evaluationInORF}
\end{equation}
the  linear velocity $\dot \xi^{L,v}$ and angular velocity $\dot \xi^{R,v}$ are with respect to the center point of the obstacle $\xi^c$. The relative position is ${\tilde \xi} = \xi-\xi^c$.

\subsection{Deforming Obstacle} \label{sec:movingSurfaces}
Obstacles and hulls can not only move, but they can also change their shape with respect to time, e.g. breathing body for a surgery robot. Conversely, the deformation of the perceived obstacle can be the result of uncertainties in real-time obstacle detection and position estimation. \\
The deformation velocity of an obstacle is evaluated as:
\begin{equation}
  \dot{\tilde{\xi}}^d = \dot{\vect \xi}^{L,d} + \dot{ \vect{\xi}}^{R,d} \times \tilde{\vect \xi}^d
\end{equation}
Where the linear velocity $\dot{\vect \xi}^{L,d}$ and angular velocity $\dot{ \vect{\xi}}^{R,d}$ are evaluated on the surface position in reference direction, and $\tilde{\vect \xi}^d = \xi - \xi^b$ is the relative position with respect to the boundary point (see Fig.~\ref{fig:dynamic_environment}). \\
The surface deformation should be explicitly given to the algorithm whenever it is known. Alternatively, it can be estimated from sensor readings, such as Lidar or camera. \\
\subsubsection{Repulsive Mode}
In many scenarios, the consideration of the obstacle's deformation is only of importance when it reduces the robot's workspace, and puts the robot at risk. It is sufficient to consider the deformation only along positive normal direction. For example for a circular object, we have:
\begin{equation}
  \dot{\tilde{\xi}}^d = \dot{\vect \xi}^{L,d} =
  \begin{cases}
    \dot r \; \vect n(\xi) \hspace{0.5cm} & \dot r > 0  \\
    0 & \text{otherwise}
  \end{cases}
\end{equation}
where $\dot {r}$ is the rate of change of the circle radius. \\

\subsection{Impenetrability with Respect to Maximum Velocity}
Many agents have a maximum velocity they can move with, further referred to as $v^{\mathrm{max}}$.
This limits the motion of obstacles which an agent can avoid to:
\begin{equation}
  v^{n} = \langle \dot{ \tilde \xi}, \vect n(\xi) \rangle < v^{\mathrm{max}}
  \qquad \text{as} \quad  \Gamma(\xi) \rightarrow 1
\label{eq:max_velocity}
\end{equation}

\begin{figure}[t]\centering
\centering
\includegraphics[width=0.6\columnwidth]{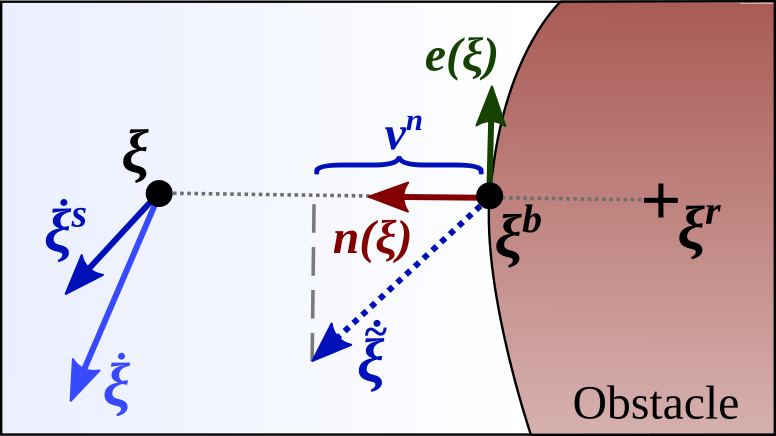}
\caption{In order to comply with a velocity limit of the robot while avoiding a collision with an obstacle of velocity $\dot{\xi}$, the modulated velocity $\dot{\xi}$ might be stretched only in normal direction to obtain the safe velocity command $\dot{\xi}^s$.}
\label{fig:dynamic_environment}
\end{figure}

When close to an obstacle, the agent must prioritize moving away from the obstacle over following the desired motion. Since the modulated velocity $\dot \xi$, see (\ref{eq:modDS_app}), does not take into account the maximal velocity, smart cropping needs to be applied. We propose the following method, which prioritizes avoidance in critical situations but sticks to the initial DS as when safe:
\begin{equation}
\dot{\xi}^{s} =
\begin{cases}
v^{n} \vect{n}(\xi) + v^{e} \; \vect{e} (\xi) & \text{if} \;\;  
\langle \frac{\dot \xi}{\| \dot{\xi}\|}, \vect{n}(\xi) \rangle  < \frac{v^{n}}{v^{\mathrm{max}}}  \\
v^{\mathrm{max}} \, \dot \xi / \| \dot \xi \| \; 
& \text{else if} \;\; \| \dot \xi \| > v^{\mathrm{max}}  \\
\dot \xi   & \text{otherwise}
\end{cases}
\label{eq:safe_velocity}
\end{equation}
with $v^{e} = \sqrt{\left(v^{\mathrm{max}}\right)^2 -\left(v^{n}\right)^2}$, see Fig.~\ref{fig:dynamic_environment}. \\ 
\\
\noindent\textbf{Theorem 3}
\textit{Consider the dynamic environment with $N^{\mathrm{obs}}$ obstacles which have a weighted local velocity $\dot{\tilde \xi}^{\mathrm{tot}}$, resulting from the obstacles' movements and surface deformations, defined in (\ref{eq:relative_obstacle_velocity}). An agent is moving in this space and has a maximum velocity of $v^{\mathrm{max}}$, further the obstacles' surface velocities are limited by (\ref{eq:max_velocity}). The agent which starts in free space, i.e. $ \Gamma_o(\{\xi \}_0) > 1, \; \forall o \in N^{\mathrm{obs}}$ and moves according to (\ref{eq:safe_velocity}), will stay in free space for infinite time, i.e. $\Gamma_o(\{ \xi\}_t) > 1, \; t = 0..\infty , \; \forall o \in N^{\mathrm{obs}}$.}   \hfill \textbf{Proof:}~see~Appendix~\ref{sec:proofTheorem3}. \\




\subsection{Reference Point Placement} \label{sec:reference_point_placement}
\subsubsection{Dynamic Extension of Hull}
\begin{figure}[t]\centering
\begin{subfigure}{.32\columnwidth} %
\centering
\includegraphics[width=\textwidth]{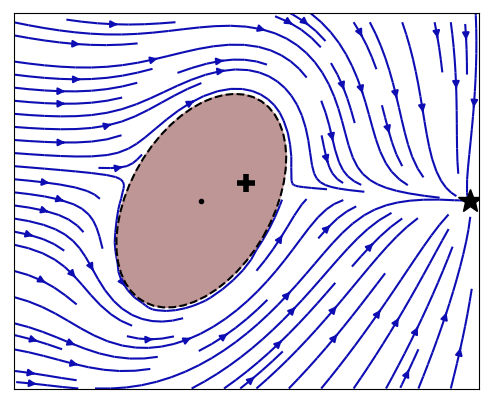}
\caption{Reference inside}
\label{fig:ellipse_dynamic1}
\vspace{0.2cm}
\end{subfigure}%
\begin{subfigure}{.32\columnwidth} %
\centering
\includegraphics[width=\textwidth]{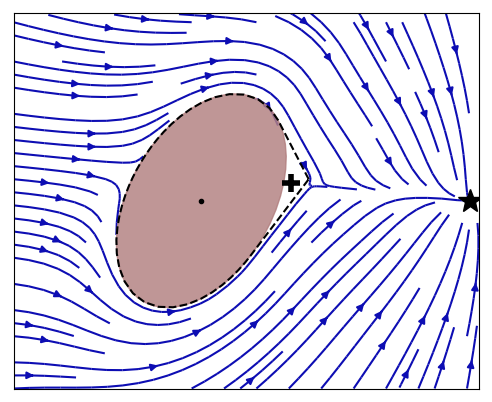}
\caption{Reference close}
\label{fig:ellipse_dynamic2}
\vspace{0.2cm}
\end{subfigure}%
\begin{subfigure}{.32\columnwidth} %
\centering
\includegraphics[width=\textwidth]{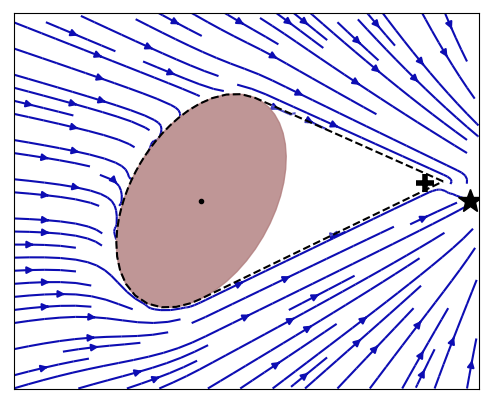}
\caption{Reference far}
\label{fig:ellipse_dynamic3}
\vspace{0.2cm}
\end{subfigure}
\begin{subfigure}{.32\columnwidth} %
\centering
\includegraphics[width=\textwidth]{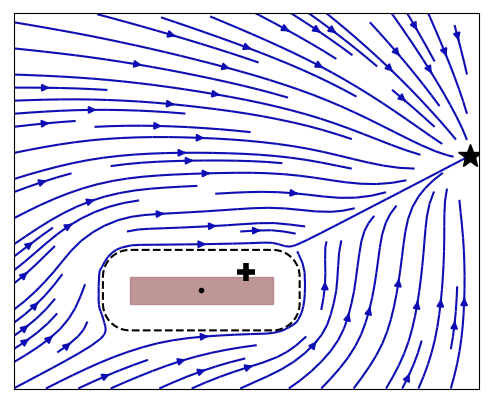}
\caption{Reference inside}
\label{fig:cuboid_dynamic1}
\end{subfigure}%
\begin{subfigure}{.32\columnwidth}
\centering
\includegraphics[width=\textwidth]{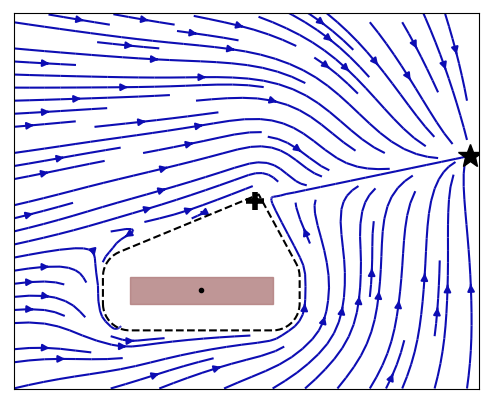}
\caption{Reference close}
\label{fig:cuboid_dynamic2}
\end{subfigure}%
\begin{subfigure}{.32\columnwidth}
\centering
\includegraphics[width=\textwidth]{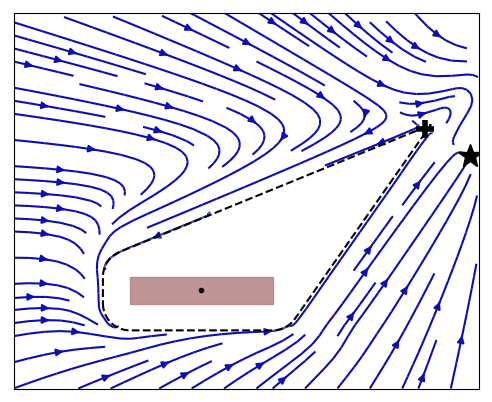}
\caption{Reference far}
\label{fig:cuboid_dynamic3}
\end{subfigure}%
\caption{Dynamic extension of the hull for an ellipsoid object without margin (a)-(c) and a nonsmooth polygon object with constant margin (d)-(e).}
\label{fig:dynamic_hull_extenion}
\end{figure}

Clusters of more than two convex obstacles do often not form a \textit{star-shape}. In such cases, we propose to extend the hull of each obstacle such that they all include a common reference point. The new hull is designed to be convex for each obstacle, and hence the cluster is star-shaped. The extended hull creates a cone that is tangent to the obstacle's surface and has the reference point at its tip (Fig.~\ref{fig:dynamic_hull_extenion}). Since the clusters are \textit{star-shapes}, there is a global convergence of the vector field towards the attractor. \\
The extension of the surfaces can be done dynamically, as a collision-free trajectory is ensured around deforming obstacles (Sec.~\ref{sec:movingSurfaces}). 

\subsubsection{Clustered Environments with Obstacles and Boundaries} \label{sec:mixed_scenario}
\begin{figure}[t]\centering
\centering
\includegraphics[width=0.99\columnwidth]{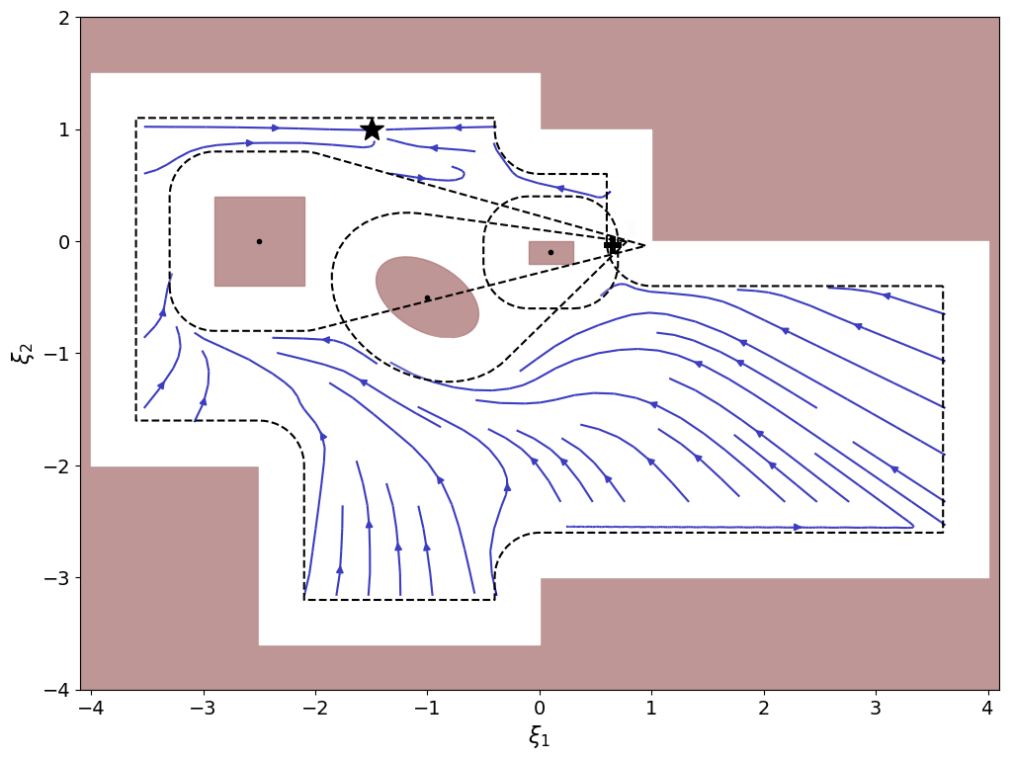}
\caption{Full convergence towards the attractor (black star) in an environment of three obstacles intersecting with the boundary.}
\label{fig:obstacle_boundary_intersection}
\end{figure}

For obstacles which intersect with the boundary, the reference point has to be placed inside the wall, i.e., $\Gamma_b(\xi^{r}_o) < 1$ (see Fig.~\ref{fig:obstacle_boundary_intersection}). This enforces all trajectories to avoid the intersecting obstacles by moving away from the wall (counterclockwise in this example). The boundary modulates them in the same direction. Hence, there is full convergence of all trajectories towards the attractor. This is true for a boundary-obstacle with a positive (local) curvature:
\begin{equation}
  c^b (\xi) > 0 \qquad \forall \, \xi
\end{equation}
with the curvature given as:
\begin{gather}
\quad c^{(\cdot)}{(\xi)} = \lim_{\Delta \xi \rightarrow \vect 0} \frac{R(\xi) - R(\xi+ \Delta \xi)}{\Delta  \xi} \nonumber \\
\quad \forall \xi \in \mathcal{X}^b, \;\; \langle \Delta \xi, \, \vect{n} (\xi)\rangle = 0 \label{eq:curvature}
\end{gather}

\editcolor{\section{Obstacle Avoidance with Robots} \label{sec:robot_arm}
The algorithm has so far been described for a point mass. It is straightforward to extend this to control robots, which can be approximated by a circle (e.g., drones, wheel-based platforms) by creating a margin around all obstacles wide enough to account for the shape of the robot. The method can also be extended to higher dimensions and multiple degrees of freedom robot arms by describing and evaluating the system (robot + obstacle) in joint-space. This, however, requires representing the obstacle in configuration space which is not always easy, especially when the obstacle moves.

Alternatively, in the rest of this section, we introduce a weighted evaluation of the desired dynamics along a robot arm’s links to obtain a collision-free trajectory towards the desired goal in Cartesian space.  

\subsection{Goal Command Towards Attractor}
Consider a robot arm with end-effector's position $\xi$. The desired velocity towards the attractor is evaluated as described in Section~\ref{sec:obstacle_avoidance} to Section~\ref{sec:dynamic_environments} and denoted as $\dot \xi$. The goal command in joint space is evaluated through inverse-kinematics as:
\begin{equation}
    \dot{\vect{q}}^g = \matr{\hat J}^{\dag}(\vect{q}) \dot{\xi}
    \label{eq:goal_command}
\end{equation}
where $\matr{\hat J}(\vect{q})$ is the Jacobian of the robot arm with respect to position only\footnote{The inverse of the Jacobian $\matr J^{\dag}(\vect q)$ is obtained through the Moore-Penrose pseudo-inverse.}.

\subsubsection{$\Gamma$-Danger Field}
The closer an obstacle is to the robot, the more it is in danger to collide. Based on the $\Gamma(\xi)$-function introduced in (\ref{eq:levelFunc}), we introduce a $\Gamma$-danger field with respect to all $N^{\mathrm{obs}}$ obstacles as
\begin{equation}
  \Gamma^d(\xi) = \min_{o \in [1 .. N^{\mathrm{obs}}]} \Gamma_o(\xi)  \label{eq:danger_gamma}
\end{equation}
The field is displayed in Fig.~\ref{fig:gamma_field_and_modulation}. This field is used to evaluate the weights which continuously from the goal command $\vect{\dot q}^g$ to the avoidance command $\vect{\dot q}^m$.

\begin{figure}[t]
    \centering
    \includegraphics[width=.9\columnwidth]{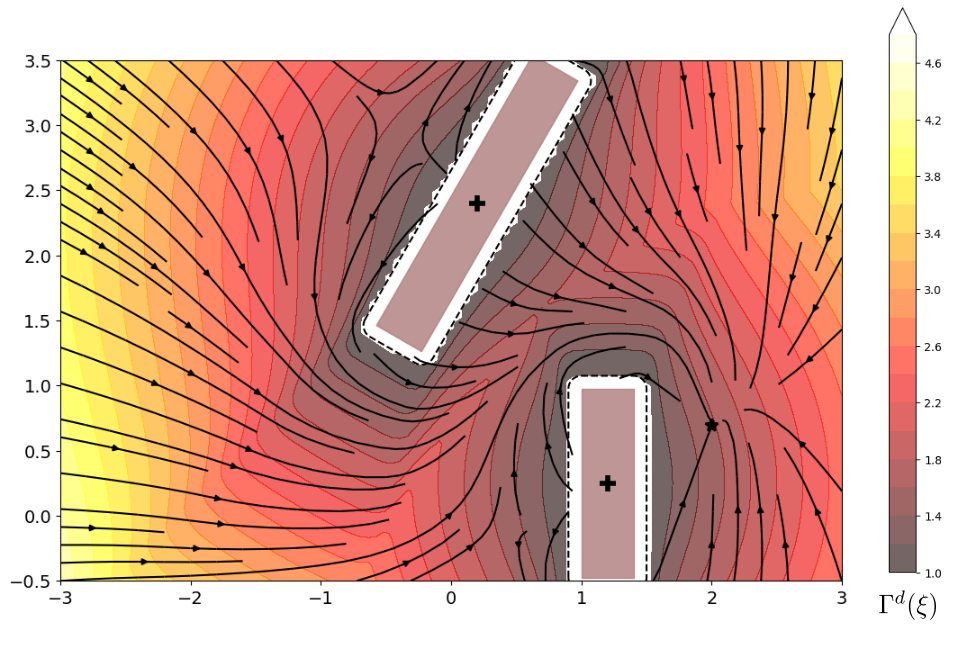}
    \caption{$\Gamma$-field and desired direction in a multi-obstacle environment.}
    \label{fig:gamma_field_and_modulation}
\end{figure}

\subsection{Link Avoidance}
The avoidance command $\vect{\dot q}^m$ ensures that each link is avoiding the collision with the environment. To extend the obstacle avoidance algorithm to a rigid body, we introduce $N^S$ section points $\xi^S_{l, s}$ for each link $l$ and $s \in [1 .. N^s]$. The dynamical system is evaluated at each section point and coupled with a section weight $w^S_s$. The section point weight $w^S_{s}$ increases the lower the $\Gamma$-danger value (see Sec.~\ref{sec:section_weights}).

\subsubsection{Rigid Body Dynamics}
The linear and angular avoidance velocity for link $l$ are obtained as:
\begin{equation}
  \vect{v}^L = \sum_{s=1}^{N^S} w^S_{s} \dot{\xi}_{l, s}^S
  \;\; \text{and} \;\;
  \omega^{L} = \sum_{s=1}^{N^S} w^S_{s} (\xi_{l, 0}^S - \xi_{l, s}^S) \times \left( \vect{v}^S_{s} - \vect{v}^L   \right)
  \label{eq:joint_avoidance_velocities}
\end{equation}
where $\xi_{l,0}^S$ is the position of the root of a link $l$, and $\dot{\xi}_{l, s}^S$ is the dynamical system based avoidance evaluated as described in in Sec.~\ref{sec:obstacle_avoidance}~-~\ref{sec:dynamic_environments}. A single link avoiding a circle can be seen in Fig.~\ref{fig:rigid_body_dynamics}.

\begin{figure}[t]
    \centering
    \includegraphics[width=.5\columnwidth]{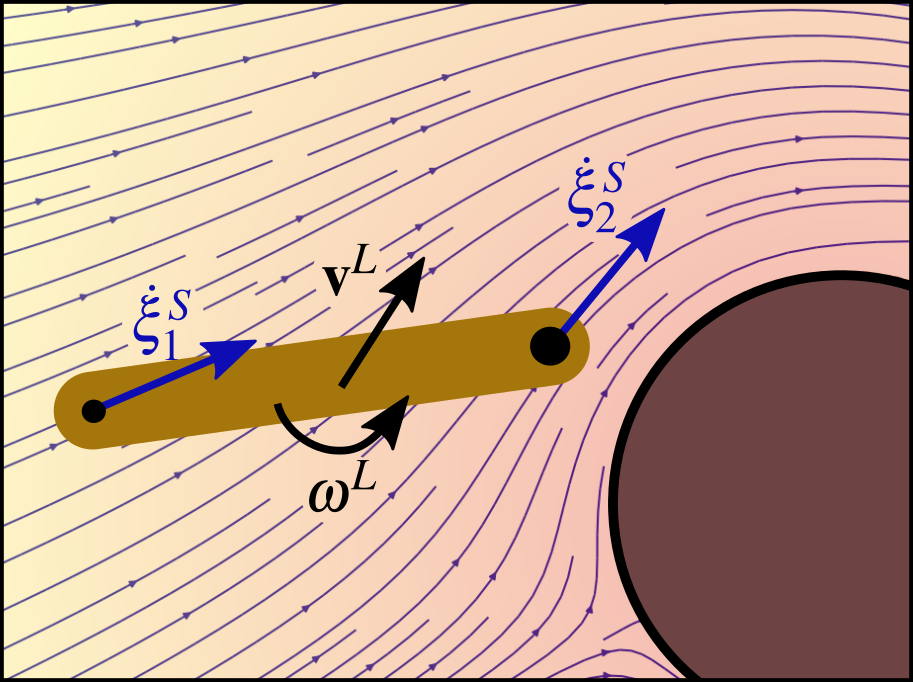}
    \caption{The resulting linear velocity $\vect v^L$ and angular velocity $\omega^L$ for a rigid body with to sections points.}
    \label{fig:rigid_body_dynamics}
\end{figure}

\subsubsection{Joint Modulation}
The desired joint avoidance command is obtained through inverse kinematics as:
\begin{equation}
    \vect{\dot{q}}^m
    = \matr{J}^{\dag}(\vect q, \xi_{l,0}^S)
    \begin{bmatrix}
    \vect{v}^L \\
    \omega^L
    \end{bmatrix}
    \label{eq:joint_modulation_command}
\end{equation}
where $\matr{J}(\vect q, \xi_{l,0}^S)$ is the Jacobian up the root of the link $l$, i.e., the positionof the section point $\xi_{l,0}^S$. 

\subsection{Evaluation Along the Robot Arm}
The evaluation is performed by iterating over all $N^L$ links, starting from to base of the robot arm. At each iteration, the joint control $\dot{\vect{q}}^c$ of all links which are part of the kinematic chain are updated, i.e., if the evaluation is performed for link $l$, the joint control are updated for $\vect{\dot q}^c_i \; \forall i \leq l$. \\
The evaluation is weighted along with the link to ensure collision avoidance of all links while trying to follow the goal command $\vect{\dot q}^g$. The link weights $w_l^L$ are further described in Sec.~\ref{sec:link_weights}.

\subsubsection{Initial Joint Control}
The joint control is initialized based on the goal control from (\ref{eq:goal_command}) as:
\begin{equation}
  \dot{\vect{q}}^c \gets \left( 1 - \sum_{l=1}^{N^L} w_l^L \right) \vect{\dot q}^g
\end{equation}

The first element of the joint command is then updated based on the avoidance velocity obtained in (\ref{eq:joint_modulation_command}) as:
\begin{equation}
  \dot{\vect{q}}^c_{[1]} \gets \dot{\vect{q}}^c_{[1]} +  w_1^L \vect{\dot q}^m_{[1]}
\end{equation}

\subsubsection{Joint Control Correction}
As an effect of influence the obstacle avoidance of each link, the obtained control command $\dot{\vect q}^c$ differs from the ideal goal command $\dot{\vect q}^g$. This difference is obtained at link $l$ as:
\begin{equation}
  \vect{v}^{\Delta} = \hat{\matr{J}}_{l}(\vect{q}) \left( \vect{\vect{\dot{q}}}^{g}_{[1:l]} - \vect{\dot{q}}^{c}_{[1:l]} \right)
  \quad \forall \; l > 1
\end{equation}
where $\vect{\dot{q}}^{c}_{[:l]}$ is the current control command, evaluated up to link $l-1$, and $\hat{\matr{J}}_{l}(\vect{q})$ is the arm Jacobian with respect to position up to link $l$.\\
The joint speed of link $l$ which would best account for this difference can be evaluated as:
\begin{equation}
  \dot q^{\Delta} = \langle {\vect{\omega}}^{\Delta},  \vect{I}^\omega \rangle
  \quad \text{with} \;\;
  {\vect{\omega}}^{\Delta} = \vect{v}^{\Delta} \times \vect {I}^L
  \label{eq:correction_control}
\end{equation}
where $\vect {I}^\omega$ is the direction of rotation of the joint actuating link $l$\footnote{We assume single-degree of freedom joints.} and $\vect {I}^L$ is the direction along with the link, i.e., pointing from one joint to the next. The variables can be observed in Fig.~\ref{fig:robot_arm_arrows}.

\subsubsection{Joint Control Update}
The velocity of each joint is evaluated one-by-one, starting at the joint closest to the base of the robot towards the end-effector (see Algorithm~\ref{alg:avoidance_robot_arm}).
The joint command $\dot{\vect{q}}^c$ is first updated by applying the correction control from (\ref{eq:correction_control}) to the current joint $l$ as:
\begin{equation}
  \dot{\vect{q}}^c_{[l]} \gets \dot{\vect{q}}^c_{[l]} + \dot{q}^{\Delta} \sum_{i=1}^{l-1} w^L_{i}
  \qquad \forall \, l > 1
\end{equation}
The modulation command $\dot{\vect{q}}^m$ from (\ref{eq:joint_modulation_command}) is then applied to all underlying joints:
\begin{equation}
  \dot{\vect{q}}^c_{[1:l]} \gets \dot{\vect{q}}^{c}_{[1:l]} + w^L_{l} \vect{\dot{q}}^m_{[1:l]}
  \qquad \forall \, l > 1
\end{equation}
This is executed iteratively for all joints $l > 1$.

\begin{algorithm}[ht]
\caption{Joint Control Command for a Robot Arm}\label{alg:avoidance_robot_arm}
\begin{algorithmic}[1]
  \renewcommand{\algorithmicrequire}{\textbf{Input:}}
  \renewcommand{\algorithmicensure}{\textbf{Output:}}
  \REQUIRE $N^L$, $N^S$, $f(\xi)$, obstacle-environment
  \ENSURE $\vect{\dot{q}}^c$
  \STATE $\xi^{r}_{o} \;\; \forall o \in {1 .. N^{\mathrm{obs}}}$ \COMMENT{update dynamic obstacles as in Sec.~\ref{sec:dynamic_environments}}
  \STATE $\dot{\vect{q}}^g \gets \left(\matr{J}(\vect{q})\right)^{\dag} \dot{\xi}$ \COMMENT{compute goal command as in (\ref{eq:goal_command})}
  \FOR{$l = 1 \textbf{ to } N^L$}
  \FOR{$s = 1 \textbf{ to } N^S$}
  \STATE $\Gamma(\xi_{s, l})$ \COMMENT{compute \textit{danger}-field as in (\ref{eq:danger_gamma})}
  \STATE $w^{\Gamma}$ \COMMENT{compute danger-weight as in (\ref{eq:gamma_weights})}
  \ENDFOR
  \STATE $w^L_l$ \COMMENT{compute link weight as in (\ref{eq:link_weight})}
  \ENDFOR
  \STATE $\dot{\vect{q}}^c \gets \left( 1 - \sum_{l} w_l^L \right) \vect{\dot q}^g$ \COMMENT{initialize control command}
  \STATE $\dot{\vect{q}}^c_{[1]} \gets \dot{\vect{q}}^c_{[1]} +  w_1^L \vect{\dot q}^m_{[1]}$
  \FOR{$l = 2 \textbf{ to } N^L$}
  \IF{$w^L_l > 0$}
  \FOR{$s = 1 \textbf{ to } N^{s}$}
  \STATE $w^S_s$ \COMMENT{compute section weight as in (\ref{eq:point_weight})}
  \ENDFOR
  \STATE $\vect{v}^L_{l}$, $\omega^L_{l}$ \COMMENT{compute avoidance velocities as in (\ref{eq:joint_avoidance_velocities})}
  \STATE $ \vect{\dot{q}}^m_{l}$ \COMMENT{compute modulation command as in (\ref{eq:joint_modulation_command})}
  \STATE $\dot{\vect{q}}^c_{[1:l]} \gets \dot{\vect{q}}^c_{[1:l]} +  w^L_{l} \vect{\dot{q}}^m_{[1:l]}$
  \ENDIF
  \STATE $\vect{v}^{\Delta} \gets \matr{J}_{ l}(\vect{q}) \left( \vect{\vect{\dot{q}}}^{g}_{[1:l]} - \vect{\dot{q}}^{c}_{[1:l]} \right)$
  \STATE ${\vect{\omega}}^{\Delta} \gets \vect{v}^{\Delta} \times \vect {I}^L$
  \STATE $\dot q^{\Delta} \gets \langle {\vect{\omega}}^{\Delta},  \vect{I}^\omega \rangle$
  \STATE $\dot{\vect{q}}^c_{[l]} \gets \dot{\vect{q}}^c_{[l]} + \dot{q}^{\Delta} \sum_{i} w^L_{i}$
  \ENDFOR
\end{algorithmic}
\end{algorithm}

\begin{figure}[ht]
    \centering
    \includegraphics[width=.6\columnwidth]{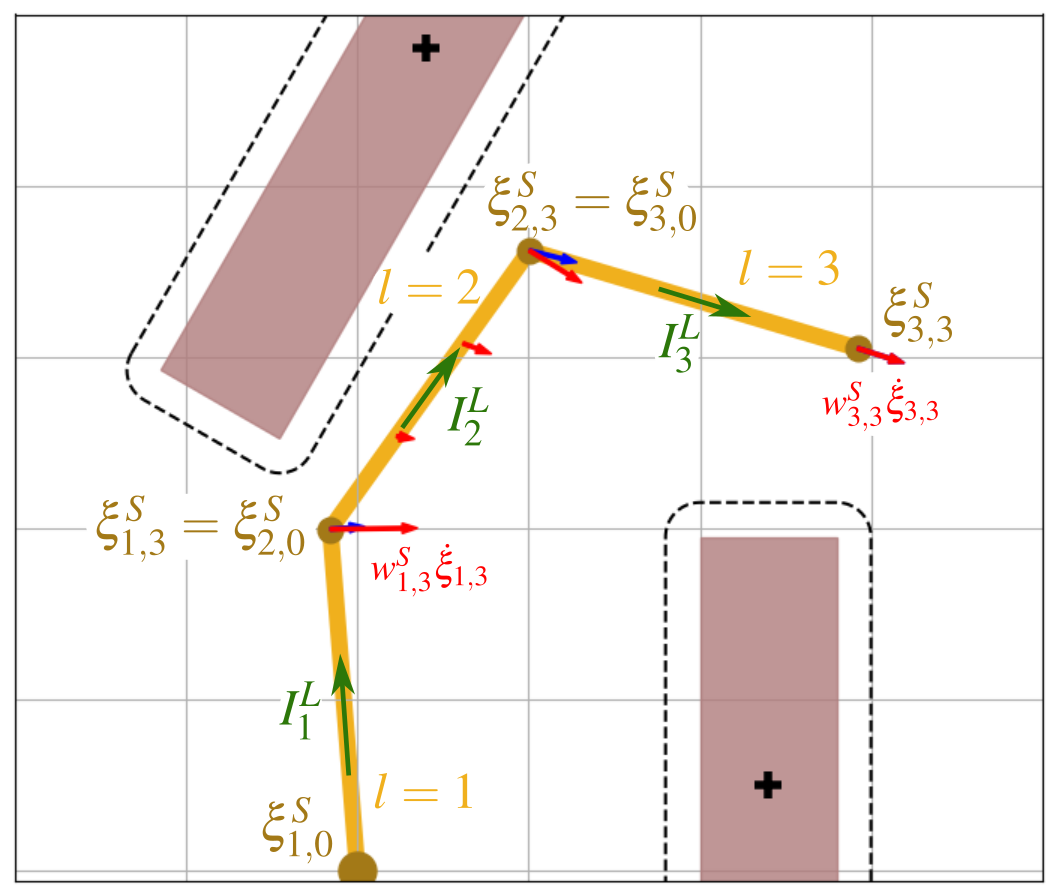}
    \caption{A robot arm with three links ($N^L = 3$), which each containing three sections points ($N^S = 3$) is surrounded by two obstacles. The blue arrows are the goal velocity obtained (only evaluated at the joints), and the red arrows are the weighted avoidance velocities. The direction of joint rotation $\vect I^\omega_l$ is pointing out of the plane.}
    \label{fig:robot_arm_arrows}
\end{figure}

\subsection{Validation in Simulation}
We applied the algorithm to two scenarios with planar robotic arms (Fig.~\ref{fig:robot_arm_avoidance}). The scenario in Fig.~\ref{fig:two_linK_robot_avoidance} was chosen similar to the one presented in \cite{zanchettin2015passivity}. Our approach can avoid obstacles in a similar setup without a navigation function. \\
The scenario in Fig.~\ref{fig:three_linK_robot_avoidance} includes a more complex robot with three links. Additionally, the skew placement of the obstacle and the position of the start end goal point require the robot to actively go around the obstacle guided by the reference point (see Fig.~\ref{fig:gamma_field_and_modulation} for the full DS).

\begin{figure}[ht]
\centering
\begin{subfigure}{.80\columnwidth} %
\centering
\includegraphics[width=\textwidth]{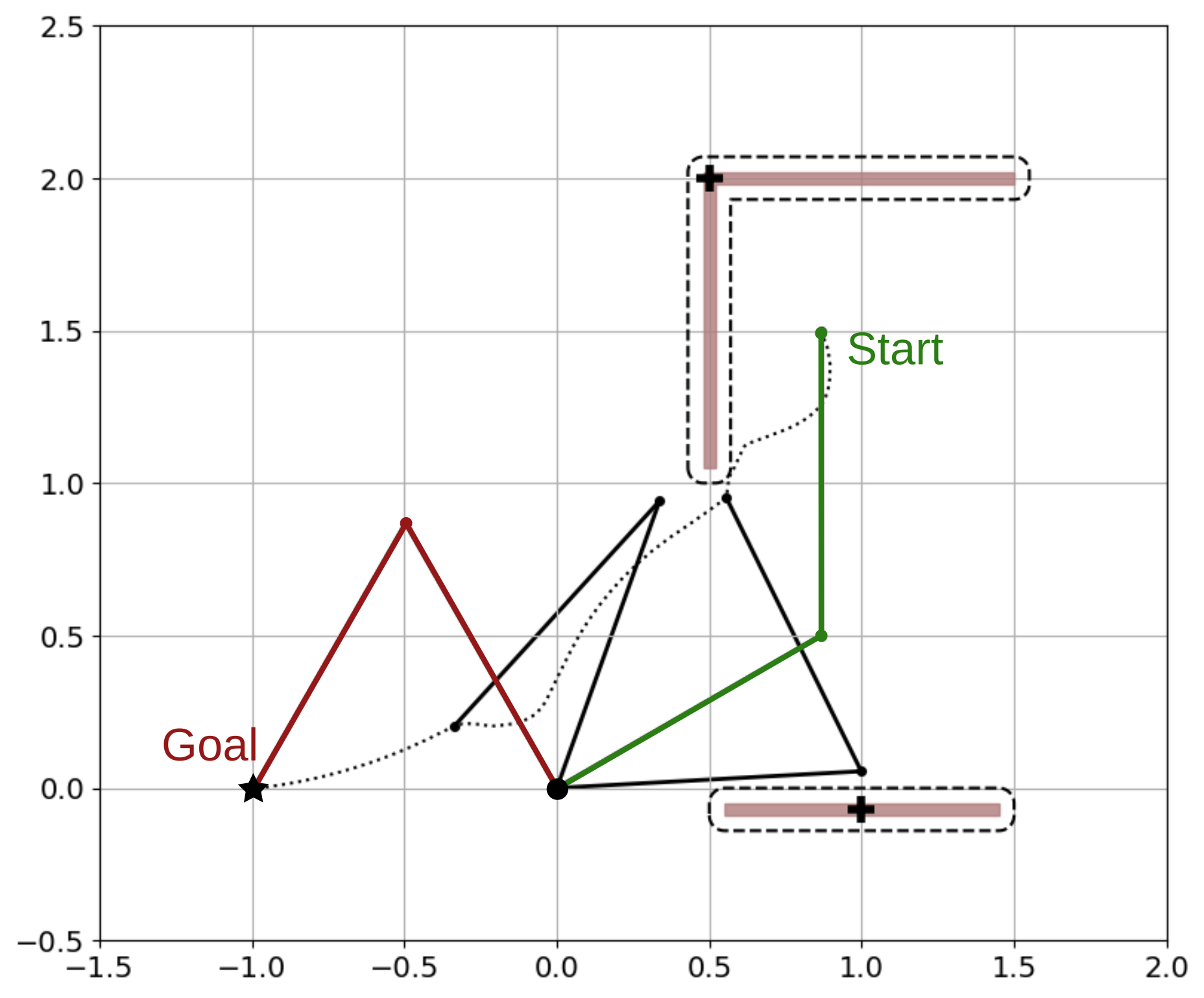}
\caption{2 DoF arm}
\label{fig:two_linK_robot_avoidance}
\end{subfigure}
\begin{subfigure}{.80\columnwidth} %
  \centering
  \includegraphics[width=1.0\textwidth]{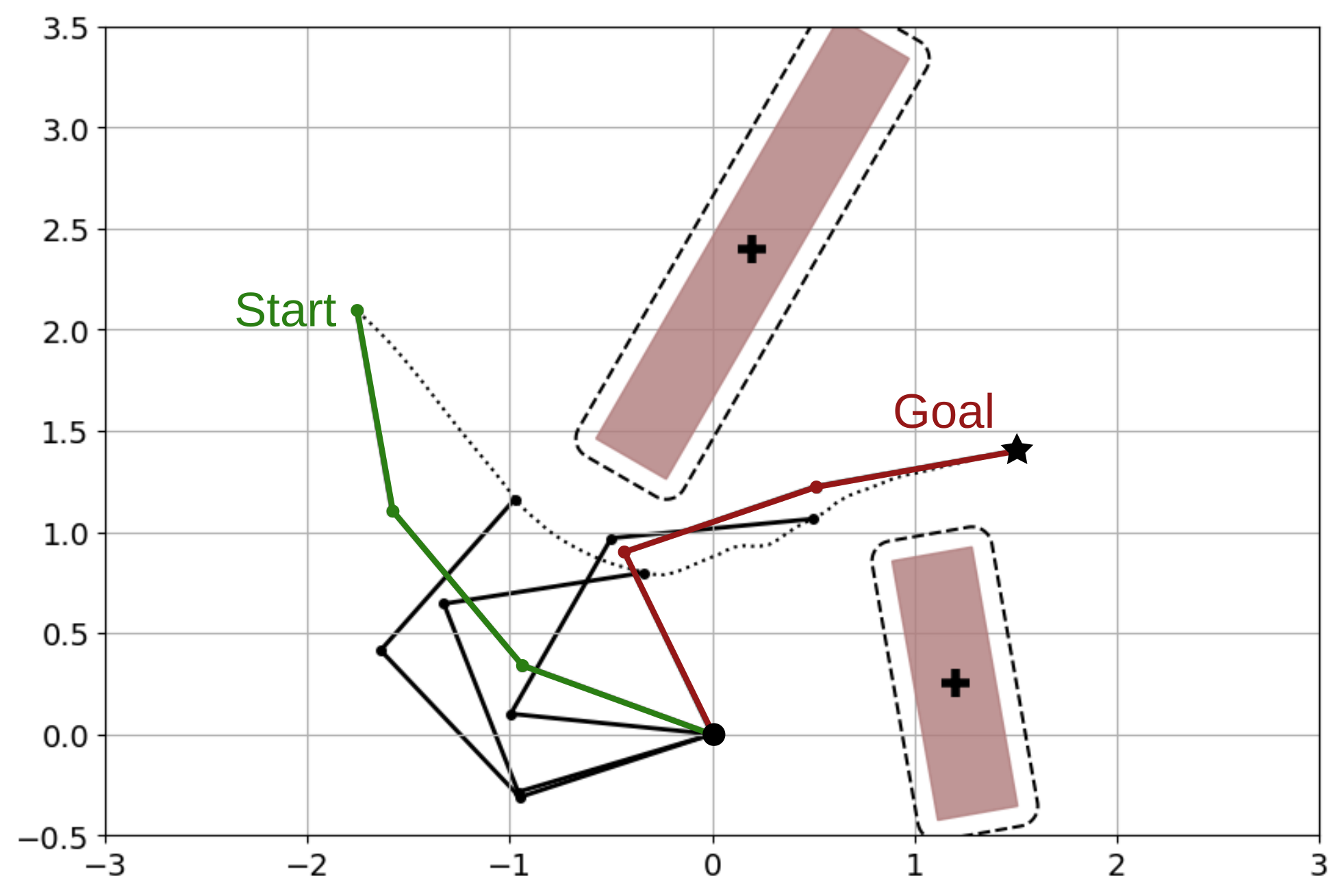}
\caption{3 DoF arm}
\label{fig:three_linK_robot_avoidance}
\end{subfigure}
\caption{Time sequence of robot arms navigating in two different environments.}
\label{fig:robot_arm_avoidance}
\end{figure}

}

\section{Comparison Algorithms}
\subsection{Qualitative Comparison}
\editcolor{We have selected multiple time-invariant, local obstacle avoidance algorithms for qualitative comparison with the presented method Table~\ref{tab:comparison_qualitative}. The \textit{Navigation Functions}, \textit{Lyapunov QP}, \textit{Sphere World QP} and \textit{Danger Fields} all rely on critical tuning parameters. This is a Lyapunov function for the \textit{Lyapunov QP}, and parameters for the diffeomorphic transformation or navigation function for the other three methods. While the parameters and functions can be obtained through manual tuning, no solution exists to set them in real-time automatically. Hence, these methods cannot be easily applied to dynamic environments. \\
Navigation functions and the diffeomorphic transform are defined globally. Their tuning parameter depends on the distribution of the obstacles all across space. Two obstacles close together far from an agent will determine the possible tuning parameters and influence the avoidance behavior. As a result, the methods cannot be transferred easily to clustered dynamic environments.\\
While other methods have already allowed navigation inside walls, this has not been done in combination with proven \textit{star-shape} world convergence and dynamic surroundings.}
\newcommand{\hCell}[1]{\parbox[b]{1.1cm}{\vspace{0.08cm} \centering \textbf{#1}}}
\newcommand{\lCell}[1]{\parbox[b]{4.0cm}{\vspace{0.05cm} \textbf{#1}}}
\begin{table*}[ht]
    \centering
    \begin{tabular}{|l|c|c|c|c|c|c|c|c|} \hline 
      & \hCell{\ \\ Dynamic} & \hCell{\\ Reference} & \hCell{\\ Orthog.} & \hCell{\\ Repulsion} & \hCell{Navigation \\ Functions} & \hCell{Lyapunov \\ QP} & \hCell{Sphere \\ World~QP} & \hCell{Danger \\ Fields} \\
      & \textit{(presented)} & \cite{huber2019avoidance} & \cite{khansari2012dynamical} & \cite{khatib1986real} & \cite{koditschek1990robot, rimon1991construction, rimon1992exact,} & \cite{reis2020control} & \cite{notomista2021safety} &  \cite{lacevic2013safety, zanchettin2015passivity} \\ \hline
      \lCell{Time invariant} & \checkmark & \checkmark & \checkmark & \checkmark & \checkmark & \checkmark & \checkmark & \checkmark \\ \hline
      \lCell{History invariant} & \checkmark & \checkmark & \checkmark & \checkmark & \checkmark & & &  \checkmark \\ \hline
      \lCell{Convergence: convex} & \checkmark & \checkmark & (\checkmark) & & \checkmark & \checkmark & \checkmark & \checkmark \\ \hline
      \lCell{Convergence: star-shape} & \checkmark & \checkmark &  &  & \checkmark & (\checkmark) & \checkmark & \checkmark \\ \hline
      \lCell{No critical tuning parameter \\ (incl. Lyapunov function)} & \checkmark & \checkmark & \checkmark & \checkmark &  & &  &  \\ \hline
      \lCell{Avoidance behavior independent \\ of global distribution} & \checkmark & \checkmark & \checkmark & \checkmark & & \checkmark & & \\ \hline
      \lCell{Closed from solution \\ (no optimization)} & \checkmark & \checkmark & \checkmark & \checkmark & \checkmark &  &  & \checkmark \\ \hline
      \lCell{Considering initial dynamics \\ (not goal position only)} &  (\checkmark) & (\checkmark) & (\checkmark) & \checkmark &  & \checkmark & \checkmark & (\checkmark) \\ \hline \hline
      \lCell{Navigation inside walls} & \checkmark & &\checkmark & \checkmark & \checkmark & \checkmark & \checkmark & \checkmark \\ \hline
      \lCell{Smooth motion around corners} & \checkmark & & & \checkmark & \checkmark & \checkmark & \checkmark & \checkmark \\ \hline
      \lCell{Clustered dynamic environment} & \checkmark & (\checkmark) & (\checkmark) & \checkmark &  &  &  & \\ \hline
    \end{tabular}
    \caption{The proposed dynamic obstacle avoidance is compared to different state-of-the-art methods.
      The last three items refer to the main contributions of this work.}
    \label{tab:comparison_qualitative}
\end{table*}

\subsection{Quantitative Comparison}
\begin{figure}[t]\centering
\centering
\includegraphics[width=0.9\columnwidth]{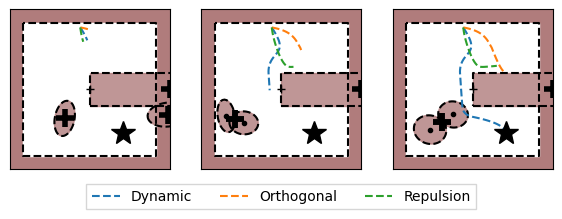}
\caption{Three snapshots of one experimental run are placed from left to right.  The obstacles are randomly initialized and move according to a random walk. One obstacle moved from the right to the left, while the other one was stationary. The three algorithms start at the same randomly chosen position and move towards the attractor.}
\label{fig:subplot_comparison}
\end{figure}

\editcolor{For a quantitative comparison, we chose algorithms that can function in clustered, dynamic environments and can handle external hulls (see Table~\ref{tab:comparison_qualitative}). \\}
The method for the modulation algorithm in dynamic environments is presented in this paper (referred as \textit{Dynamic} during this section). It is compared to \cite{khansari2012dynamical}, which uses modulation matrix based on an orthogonal decomposition matrix $\matr{E}(\xi)$ (referred as \textit{Orthogonal}) and the potential field algorithm \cite{khatib1986real} (referred to as \textit{Repulsion}). \\
The comparison is made in a simulated environment (Fig.~\ref{fig:subplot_comparison}). Two ellipse-shaped obstacles randomly change shape, and the movement is inspired by \textit{random walk}. The combined maximum expansion and obstacle's velocity are lower than the maximum speed of the agents of $ v^{\mathrm{max}} = 1\, \si{\m / \s}$. The three algorithms are given the same attractor as a goal. They start at the same time and encounter the same environment. \\
The \textit{Dynamic} algorithm is observed to have the highest rate of convergence (Tab.~\ref{tab:comparicson_convergence}) resulting from the increased environment information through the reference point. The \textit{Repulsion} has a preferable behavior on avoiding collisions because of its conservative behavior around obstacles (moving far away and only approaching them slowly). This influences the distance traveled and the time needed to reach a goal (Tab.~\ref{tab:comparison_metrics}). The mean of the velocity is lower for the \textit{Dynamic} algorithm. This is a result of no \textit{tail-effect} behind the obstacles (see Sec.~\ref{sec:repulsive_eigenvalue}). The variance of the velocity is similar for the three algorithms.\footnote{The source code of the implementation is available on \url{https://github.com/epfl-lasa/dynamic_obstacle_avoidance}}
\begin{table}[t]
    \centering
    \begin{tabular}{|l|c|c|c|} \hline
    & Converged & Collided & Minimum \\ \hline
    Dynamic & 77\% & 23\% & 0\% \\
    Orthogonal & 20\% & 23\% & 57\% \\
    Repulsion & 39\% & 1\% & 60\% \\  
    \hline
    \end{tabular}
    \caption{The outcome of the 300 trials runs were either full convergence, collision with an obstacle or getting stuck in a local minimum.}
    \label{tab:comparicson_convergence}
\end{table}

\begin{table}[t]
    \centering
    \begin{tabular}{|l|c|c|c|c|} \hline
    & d $[m]$ & t $[s]$ & $\bar v$ $[m/s]$ & $\sigma_{v}$ $[m/s]$ \\ \hline
   Dynamic & 9.69 $\pm$ 1.19 & 1.03 $\pm$ 0.13 & 0.61 $\pm$ 0.05 & 0.34 $\pm$ 0.02 \\
   Orthogonal & 10.06 $\pm$ 1.53 & 1.17 $\pm$ 0.21 & 0.55 $\pm$ 0.05 & 0.34 $\pm$ 0.02 \\
   Repulsion & 9.8 $\pm$ 1.29 & 1.39 $\pm$ 0.2 & 0.46 $\pm$ 0.03 & 0.21 $\pm$ 0.02 \\
     \hline
    \end{tabular}
    \caption{The mean and the standard deviation (after the $\pm$) are compared for the three algorithms from the 54 trials where all three agents converged. The metrics of  distance (d), duration of the run (t), the mean velocity ($\bar v$) and the standard deviation of the velocity ($\sigma_v$) are listed.}
    \label{tab:comparison_metrics}
\end{table}

\editcolor{\subsection{Computational Complexity}
The presented algorithm is closed-form, whereas the \textit{matrix inverse} is the most complex computation. Since it is applied to all obstacles, the complexity follows as $\mathcal{O}(d^{2.4} N^{\mathrm{obs}})$, a function of the number of dimensions $d$ and the number of obstacles $N^{\mathrm{obs}}$. It had an average time of \SI{3.48}{ms} on a computer with \textit{8 Intel Core i7-6700 CPU @ 3.40GHz}.\\
This is more complex than the \textit{Orthgonal}, where the matrix multiplication is the most complex operation. The complexity follows as $\mathcal{O}(d^2 N^{\mathrm{obs} })$. This reflects in the slightly faster evaluation time of \SI{2.84}{ms}. \\
The potential field is the simplest of the three algorithms, with the norm being the most complex operation and a complexity $\mathcal{O}(d N^{\mathrm{obs}})$. It has the lowest evaluation time of \SI{1.75}{ms}. \\
 The search for the optimal reference point is the computationally most extensive calculation because it requires the (iterative) closest-distance evaluation between obstacles. The Python library \textit{shapely}\footnote{https://github.com/Toblerity/Shapely} was used for the implementation, and an evaluation time of \SI{5.14}{ms} was obtained.}

\section{Empirical Validation} \label{sec:empricvalValidation}
\begin{figure}[t]
    \centering
    \includegraphics[width=0.7\columnwidth]{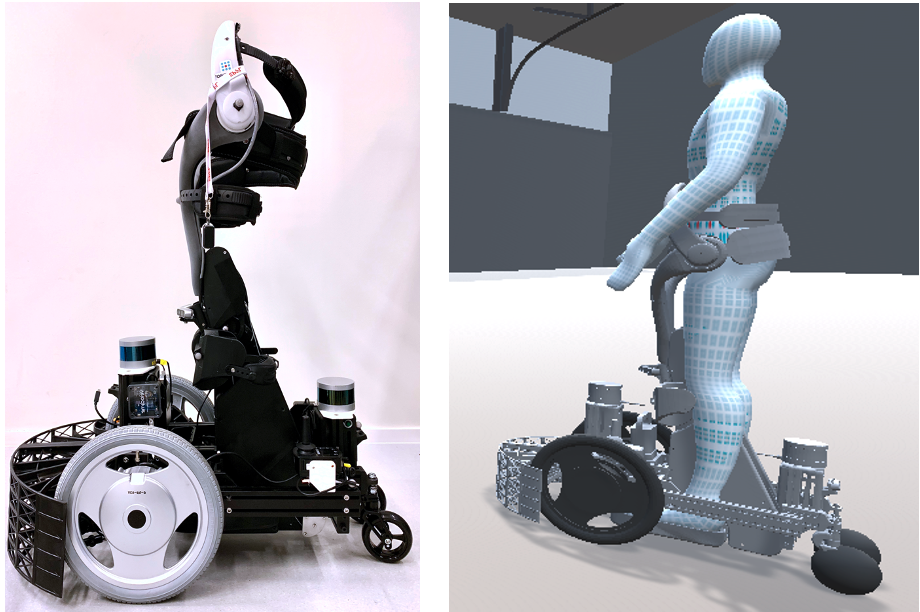}
    \caption{A picture of the semi-autonomous wheelchair real (left)
      and the simulation rendering including an operator (right).}
    \label{fig:qolo}
\end{figure}

The empirical validation is performed with the mobile robot \textit{QOLO} \cite{granados2018unpowered}, see Fig.~\ref{fig:qolo}; first in simulation and then on a real robot platform. QOLO, a semi-autonomous wheelchair, is designed to navigate in pedestrian environments and indoors. This platform is hence suited to test our algorithm's ability to avoid moving obstacles (pedestrians) and non-convex obstacles (walls, indoor furniture) containing sharp edges (tables, shelves).
In all our experiments, we assume that QOLO has information about the goal, i.e., the attractor $\xi^a$ of our nominal DS. A video of the experiments is available at \footnote{www.To Be Uploaded.com}.

\subsection{Static Environment}
\begin{figure}[t]\centering
\begin{subfigure}{.48\columnwidth} %
\centering
\includegraphics[width=\textwidth]{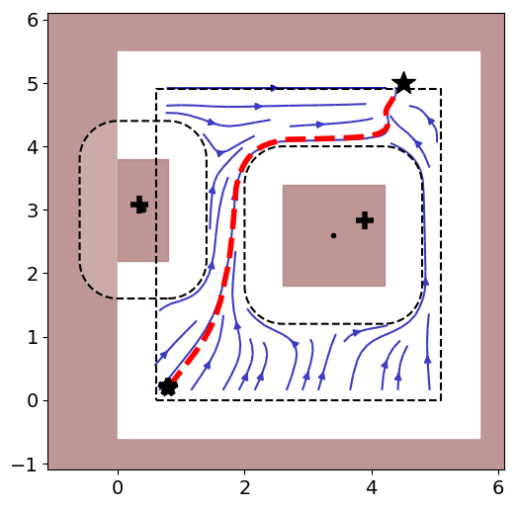}
\caption{Office with two passages}
\label{fig:qolo_static_two_tables}
\end{subfigure}%
\begin{subfigure}{.48\columnwidth} %
\centering
\includegraphics[width=\textwidth]{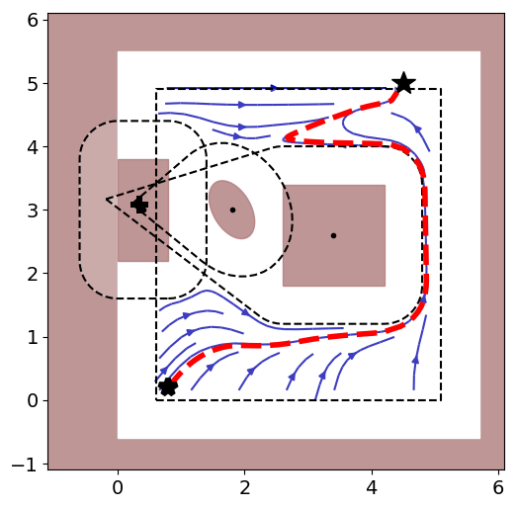}
\caption{Office with passage at side}
\label{fig:qolo_static_puppet}
\end{subfigure}%
\caption{Two static office environments with two tables in a rectangular room. The center table divides the room into two passages in (a). In (b) one of the passages is blocked by a static person. Convex expansion of the hull around the reference point (black cross) ensures convergence towards the attractor (black star). The path followed by the agent is visualized in red.}
\label{fig:qolo_static}
\end{figure}

We task QOLO to navigate in an office-like environment. The room is a square (\SI{5}{m} x \SI{5}{m}), modeled as a boundary obstacle. Further, two tables are located in the room, one at the side and one at the center. The robot starts from the bottom left, and the attractor $\xi^a$ is placed at the opposite side of the room (illustrated with a star). All objects, including the wall, are static and known a priori, and the localization is performed using the SLAM algorithm. The robot evaluates the modulated avoidance in real-time. We run the following two scenarios:
\begin{enumerate}
\item QOLO is in the room, and there are two possible paths to go around the center table. The dynamical system is split by the obstacle at the center (Fig.~\ref{fig:qolo_static_two_tables}). The robot chooses its preferred trajectory at runtime. \\
\item Additionally there is a (static) person in the room, which blocks the center passage. \\
The reference point of the obstacles is automatically placed inside the wall. The robot finds its path around the obstacles (Fig.~\ref{fig:qolo_static_puppet}). 
\end{enumerate}

\subsection{Dense Crowd (Simulation)}
\begin{figure}[t]\centering
\begin{subfigure}{.48\columnwidth} %
\centering
\includegraphics[width=\textwidth]{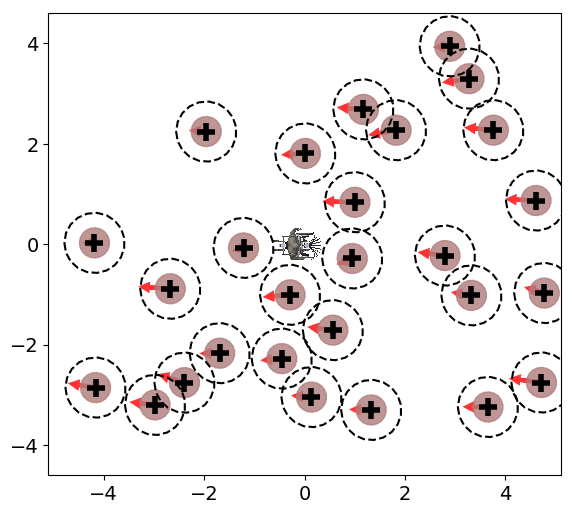}
\caption{All crowd agents}
\label{fig:qolo_in_dense_crowd_all_agents}
\end{subfigure}%
\begin{subfigure}{.48\columnwidth} %
\centering
\includegraphics[width=\textwidth]{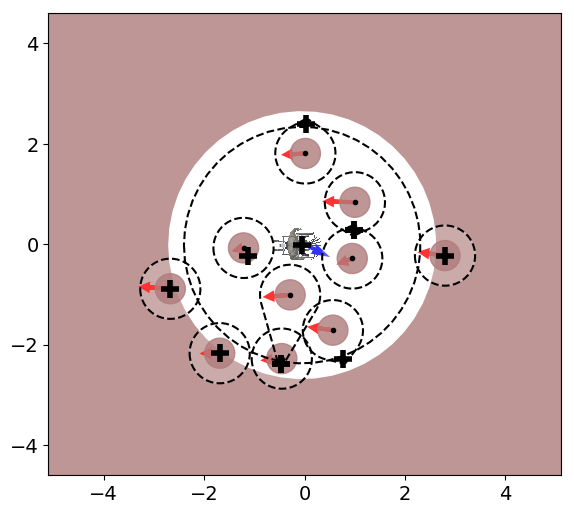}
\caption{Local Obstacle Hull}
\label{fig:qolo_in_dense_crowd_simplified}
\end{subfigure}%
\caption{The environment with many agents (left) is reduced to a scenario with 10 obstacles and an enclosing hull (right).}
\label{fig:qolo_in_dense_crowd}
\end{figure}

The robot is navigating in a corridor within a dense, simulated crowd. The motion of the crowd is created according to \cite{grzeskowiak2020toward}. Two hundred people are moving uniformly along the \SI{6}{m} wide corridor, either in the same or opposite direction as the robot. \\
QOLO is tasked to travel from one end of the corridor to the other. It is guided by the attractor of the nominal DS. All pedestrians are modeled as circular obstacles with radius of \SI{0.6}{m} (Fig.~\ref{fig:qolo_in_dense_crowd_all_agents}).\\
At each timestep, the problem is reduced to avoiding a subset of the pedestrians. Due to the crowd's density, the robot could realistically perceive only a subset of the pedestrians in real-time. The number of perceived people is set to $N^{c}=10$. The rest of the people is hidden behind a virtual, circular wall. The center of the circular wall $\xi^{c,w}$ is displaced from the position of the robot $\xi^Q$ based on the remaining obstacles:
\begin{equation}
\xi^{c,w} = \xi^Q + \sum_{i=N^c+1}^{N^{\mathrm{obs}}} \frac{\xi^{c}_i - \xi^Q}{\|\xi^{c}_i - \xi^Q\|} e^{-(\|\xi^{c}_i - \xi^Q\| - r^{p} - r^{Q})}
\end{equation}
where $i$ is iterating over the list of the obstacle which are ordered based on their distance to the robot. The displacement factor is with respect to the radius of each pedestrian, $r^p =$\SI{0.6}{\metre}, and the robot radius, $r^Q=$\SI{0.5}{\metre}.\footnote{For a real-world implementation sensory distance measurements in the horizontal plane can be used to create the virtual circular wall and its displacement, since the detection of people is still a time intensive task.} \\
The radius of the hull is chosen such that the next closest obstacle $N^c+1$ is fully within the hull. The resulting environment has a dynamic hull with changing center-position and radius (Fig.~\ref{fig:qolo_in_dense_crowd}).\\
Reducing the environment to only sphere obstacles decreases the computational time since there is a closed-form solution for the closest distance between two spheres. The evaluation on ROS1\footnote{https://www.ros.org/} and Python~2.7 run at around \SI{200}{Hz} on a \textit{Up Board: Intel Celeron N3350} with \SI{2.4}{GHz} (CPU) and \SI{8}{GB} of RAM. The number of nearby obstacles(including the wall) was eleven, while the wall remained far from the agent, it helped guide the robot around the local crowd (see Sec.~\ref{sec:mixed_scenario}). This is done by placing a reference point inside the wall if a crowd-cluster is touching the wall (see small cluster at the bottom in Fig.~\ref{fig:qolo_in_dense_crowd_simplified}). Further, fast contraction of the boundary can happen when the local crowd density is high. This forces the obstacle to stay away from surrounding obstacles.
\begin{figure}
\centering
\includegraphics[width=0.8\columnwidth]{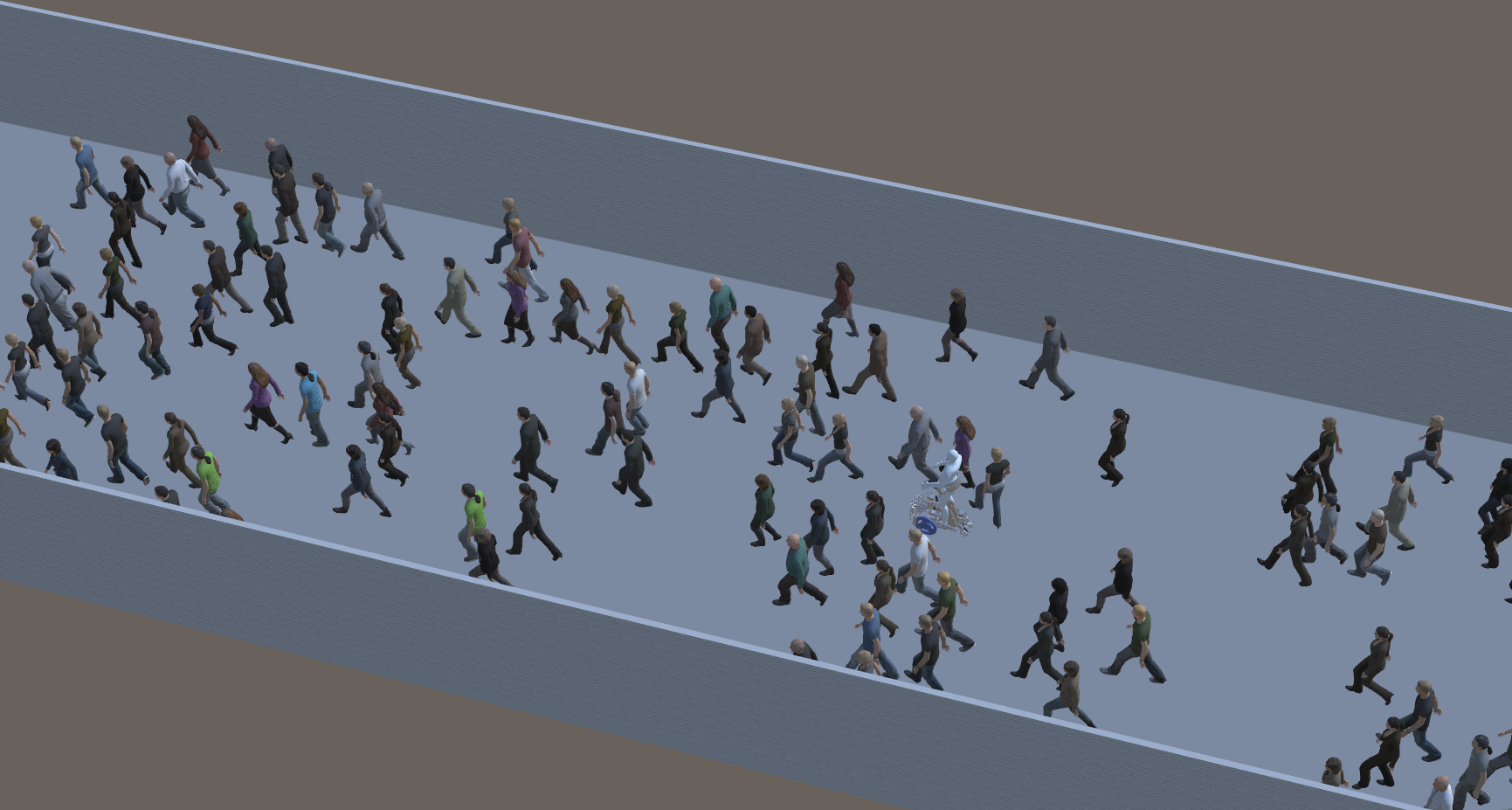}
\caption{QOLO moving in a crowd.}
\label{fig:simulation_extraction}
\end{figure}

\subsubsection{Quantitative Analysis}
\begin{figure}
\centering
\includegraphics[width=1.0\columnwidth]{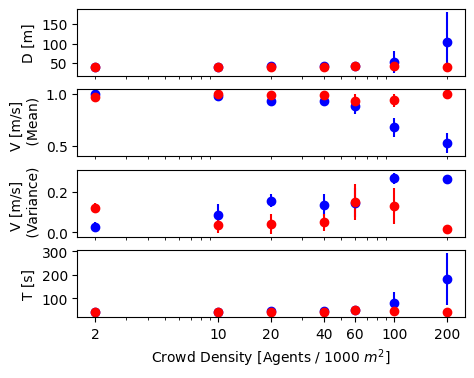}
\caption{The QOLO agent is moving in parallel (red) and opposite direction (blue) to the crowd. When the robot moves with the crowd, the density of the crowd has a negligible effect. When the robot moves in the opposite direction, the denser the crowd, the larger the cumulative distance (D) and mean velocity (V), as well as time (T), needed to reach the end of the corridor. The standard deviation of the velocity (V Variance) increases for dense crowds with counterflow}
\label{fig:qolo_simulation_graph}
\end{figure}

We evaluate the effect of the crowd size on the time it takes for the robot to travel through the corridor. The crowd moves along the (infinite) corridor at an average velocity of \SI{1}{\metre\per\second}. The simulation runs with a steady-state crowd-flow. QOLO is tasked to move in the same or opposing direction crowd's flow with the desired velocity of \SI{1}{\metre\per\second}.  \\
We assess the time, speed, and distance traveled by the robot when moving with and opposite to the flow, see Fig.~\ref{fig:qolo_simulation_graph}. When moving with the flow (parallel-flow), the crowd has no significant effect on the distance traveled by the agent or the velocity. When moving against the flow of the crowd, a decrease of the robot's velocity can be observed for crowds denser than 20 agents per 1000 square meters. (No effect on the crowd was observed since only a single robot was moving against a large crowd.) The distance traveled increases significantly for densities above 100 agents per 1000 square meters. As a result, the average time needed to reach the goal more than doubles for a crowd size of 100 people compared to 2 people. \\
In counterflow scenarios, the standard deviation of the flow increases. This results from situations where the robot has to slow down or stop to avoid the upcoming agents.

\subsection{Proof of concept: Outdoor Environment}
\begin{figure}
\centering
\includegraphics[width=0.8\columnwidth]{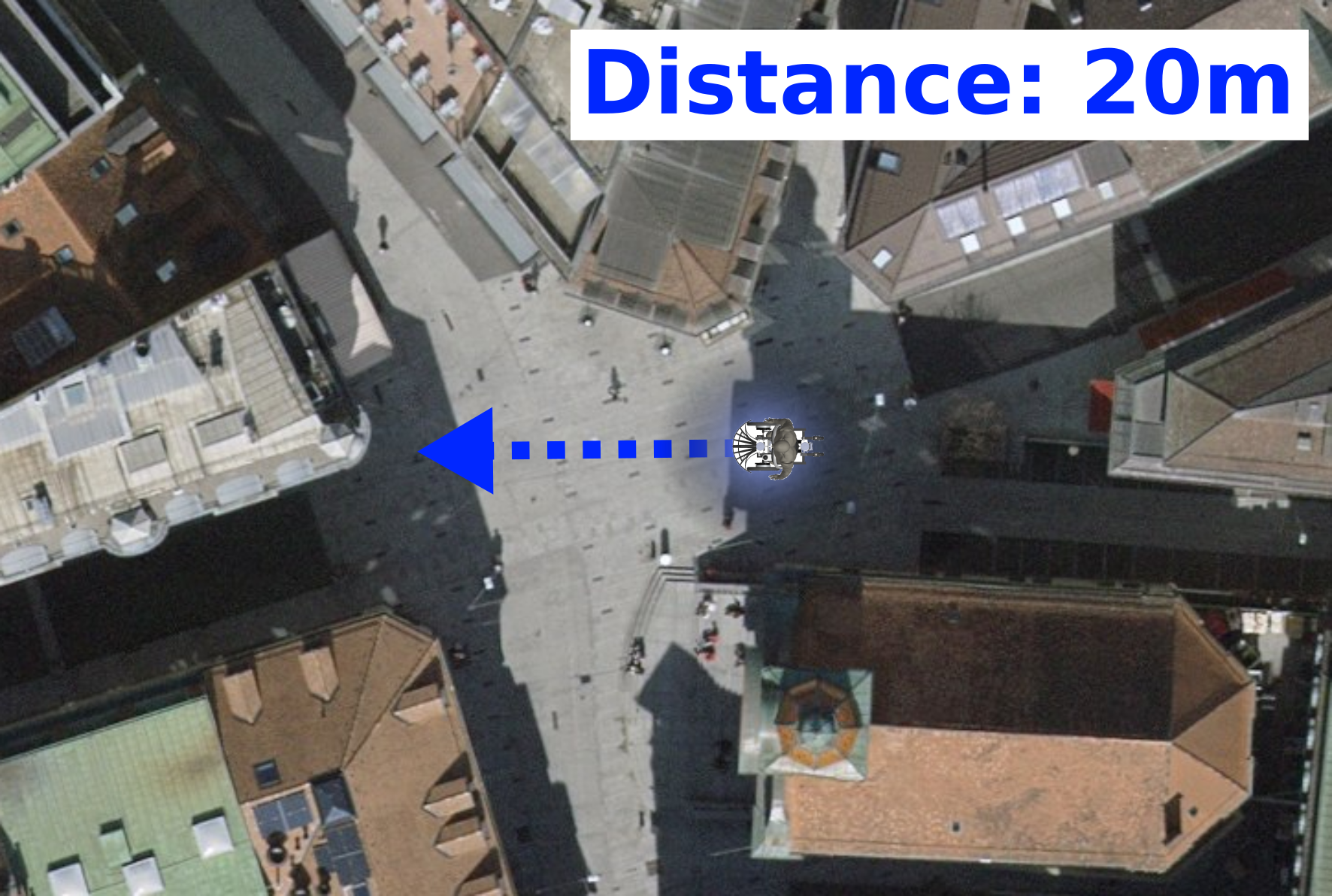}
\caption{The desired path of the robot in the outdoor environment is the direct line from the initial position of the robot (right) to the target position on the left.}
\label{fig:outdoor_map}
\end{figure}

\begin{figure}[t]\centering
\begin{subfigure}{.46\columnwidth} %
\centering
\includegraphics[width=\textwidth]{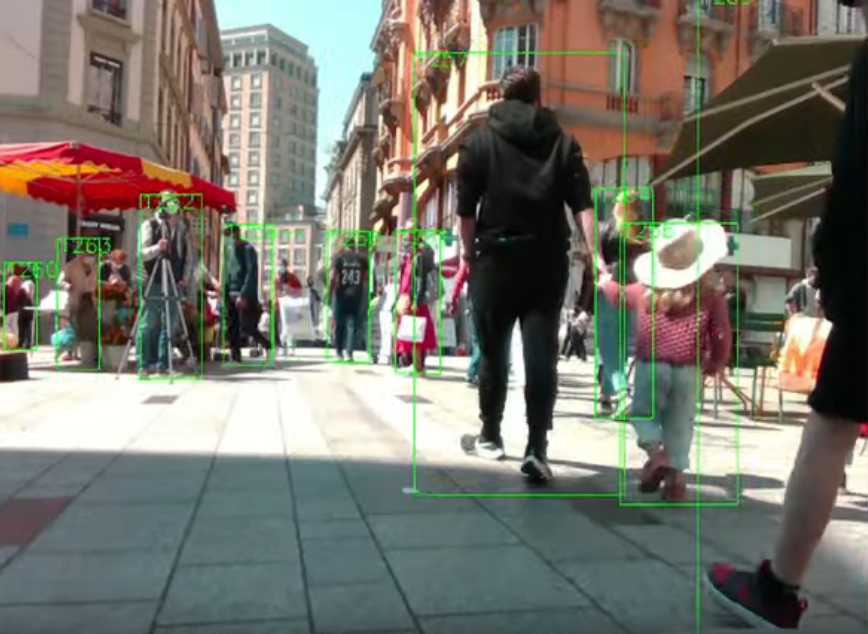}
\caption{Camera view}
\label{fig:qolo_camera}
\end{subfigure}%
\begin{subfigure}{.50\columnwidth} %
\centering
\includegraphics[width=\textwidth]{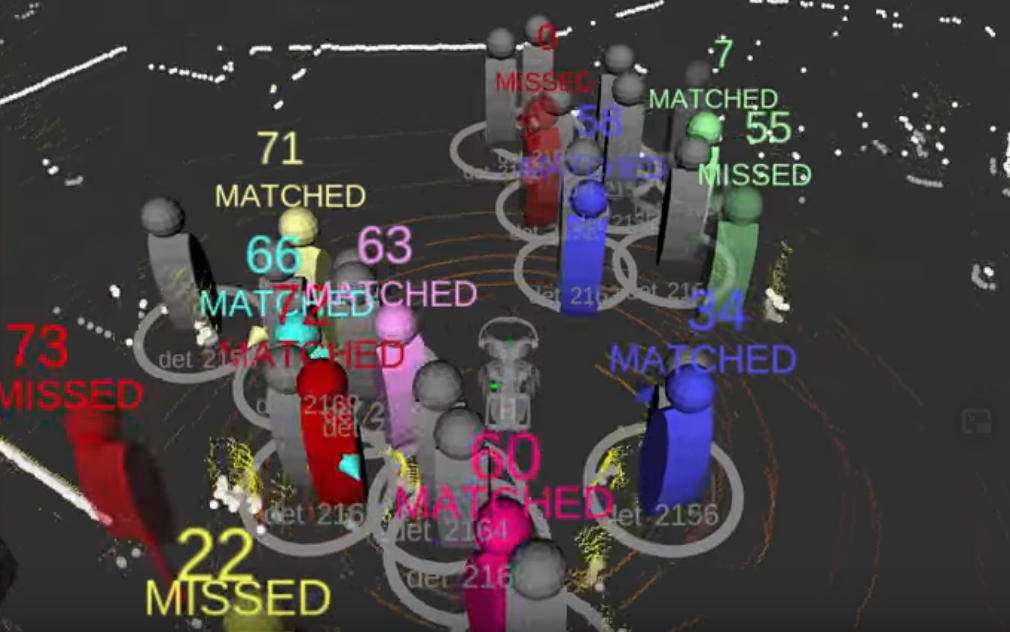}
\caption{Detector interpretation}
\label{fig:qolo_tracker}
\end{subfigure}%
\caption{The camera (a) and the LIDAR of the robot are interpreted by the detector (b), which is used for the obstacle avoidance algorithm.}
\label{fig:qolo_camera_tracker}
\end{figure}

\begin{figure}[t]
\centering
\includegraphics[width=0.8\columnwidth]{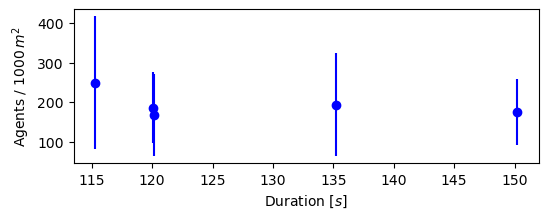}
\caption{The crowd-density is highly varying during the five runs.}
\label{fig:outdoor_evaluation}
\end{figure}

A qualitative proof of concept was performed in an outdoor environment. We brought the QOLO robot to the center of Lausanne, Switzerland city\footnote{Approval was obtained from the EPFL Ethics board and the police of Lausanne city. A driver was on board the robot during the experiment. He could start and stop at all times. A second experimenter was watching the scene and verified the output of the tracker. This was necessary in case the detector/tracker dis-functioned.}. The robot was tasked to travel back and forth across a small marketplace (Fig.~\ref{fig:outdoor_map}). The location is restricted to pedestrians only, and a total of six streets meet at the crossing. This results in a large diversity in both the pedestrians' speed and direction of movement. The robot's controller is initialized with a linear DS to reach a goal \SI{20}{m} away from the onset position. Pedestrians are detected with a camera and Lidar-based tracker developed by the authors of \cite{jia2020dr}. The output of the tracker is displayed in Fig.~\ref{fig:qolo_camera_tracker}. Recordings were taken on Saturday morning when the market was ongoing, and the crowd had a high density. \\
The non-holonomic constraints of QOLO are taken into account by evaluating the dynamical system \SI{0.53}{m} in front of the center of the wheel-axes. The linear command of the robot is the velocity part in the moving direction. The angular velocity follows the perpendicular part of the velocity. These velocities are provided to the low-level controller of the robot. \\
The geometry of QOLO is taken into account by placing a margin of \SI{0.5}{m} around each pedestrian. \\
A total of five runs were executed. The robot reached its goal autonomously without intervention. The driver reported \textit{high angular acceleration} during parts of the trip. \\
Post-hoc analysis of the video recordings revealed that the crowd density varied with a mean between 150 and 260 people per 1000 square meters (Fig.~\ref{fig:outdoor_evaluation}). The time to complete the runs ranged from \SIrange{115}{150}{s}. No correlation was observed between the density of the crowd and the time taken to reach the goal. We expect this to result from external factors influencing the run, such as the speed and direction of the crowd. \\
We see this as a successful proof-of-concept of the obstacle avoidance algorithm in a real crowd scenario. The crowd motion was more complex than the streamline simulation, as people would come from all directions and would not group in a uniform flow. Moreover, the crowd included diverse pedestrians, from families with small children to elderly people.

\section{Discussion} \label{sec:discussion}
In this work, we have introduced a Dynamical System-based obstacle avoidance algorithm. The modulation-based approach has a theoretical proof of convergence and can be applied to higher-dimensional space. The implementations presented in this work focus on collision avoidance in two dimensions, such as navigation of mobile robots and obstacle avoidance for simplified robot arms. \\
The inverted obstacle has shown to be a great representation of workspace boundaries of mobile robots, such as walls of a room or the local window in dense crowd navigation. These boundaries could be interpreted as control constraints similar to control barrier functions (CBFs). Recent works have used such control barrier function in the context of safe learning by demonstration \cite{ames2019control, taylor2020learning} or reinforcement learning \cite{cheng2019end}. Other than existing methods, we propose a closed-form solution for star-shaped barriers. \\
Since many human-made environments contain nonsmooth surfaces (e.g., tables, corners of rooms), the solution for providing a smooth flow without extending the hull has shown suitable for practical implementations. \\
The concept of dividing dynamical systems into direction and magnitude and the presented method summing vectors to avoid local minima has been used throughout: (1) to create a smooth pseudo-normal for polygon obstacles and walls and from that a smooth flow in their presence, (2) to sum the flow created by several obstacles without creating local minima, and (3) to solve the optimization problem to find the closest distance for pairs of obstacles.  \\
Finally, the algorithm has been successfully applied to a non-holonomic robot in a static indoor environment and dynamic outdoor crowds.
\editcolor{ \subsection{Scalability and Speed}
The implementation used for the experiments on QOLO was running on Python~2.7. Note that already the update to Python~\textgreater 3.6 gives around a 1.5-speed improvement. Additionally, switching to a compiled language like C++ can increase the speed by a factor of around 10 \cite{fourment2008comparison}.  We estimated that our algorithm could run at a frequency of \textgreater \SI{1}{kHz} onboard a mobile robot. This is well above the human-reaction time of around \SI{250}{ms}. \\
The bottleneck of the current implementation is the image recognition/tracking since it was only able to run at an average frequency of around \SI{5}{Hz}. This is due to the computationally heavy evaluation of the deep-neural networks used for obstacle recognition. Nevertheless, we believe that the approach of separating perception and motion-planning is favorable, supported by current trends in self-driving cars and autonomous vehicles \cite{badue2021self, schwarting2018planning}.}

\editcolor{\subsection{Contribution}
The proposed method adds value to the current state-of-the-art, not for global path planning but for reactive obstacle avoidance with convergence proofs. The modulation based avoidance algorithm fully converges in (local) \textit{star-shaped} environments towards the desired attractor $\xi^a$. The behavior is similar to approaches using potential artificial fields. However, the presented approach does not require finding and deriving an artificial potential function, but the modulation directly outputs the desired velocity.\\
Compared to other closed-form and QP-based obstacle avoidance algorithms, our method does not require any tuning of critical parameters for its convergence, nor the finding of a Lyapunov candidate function. All parameters presented in this paper can be chosen within the defined range, and theoretical convergence is ensured. \\
In the presented paper (and initially introduced in \cite{huber2019avoidance}), we presented several methods of how to place the reference point (i.e., tune its position). Even though its position is critical for convergence, it is known for \textit{star-worlds}:
\begin{itemize}
\item \textbf{Convex Obstacles:} The reference point can be placed anywhere within the obstacle, i.e. $\xi^r \in \mathcal{X}^i$.
\item \textbf{Star-shapes:} The reference point has to be placed within the kernel of the obstacle. Most of the time, this is just the geometrical center of the obstacle. For polygons, an algorithm to find the kernel has been described in \cite{lee1979optimal}.
\item \textbf{Intersecting Convex Obstacles} If several convex obstacles intersect and form a \textit{star-shape}, the reference point can be placed anywhere at the center of all intersecting obstacles $o$, i.e., $\xi^r \in \mathcal{X}^i_o \;\; \forall o$. \\
If the convex obstacles do not form a \textit{star-shape} or are intersecting with the boundary, the hull can be extended dynamically, as described in Sec.~\ref{sec:reference_point_placement}.
\end{itemize}
The placement of the reference point only becomes challenging when obstacles merge or separate dynamically. While we propose approximations for many cases (see Sec.~\ref{sec:gradient_descent} and Sec.~\ref{sec:intersecting_obstacle_descent}), we do not provide a solution for all scenarios. \\
To the best of the authors' knowledge,  there currently exists no closed-form method to generate a flow around merging and dividing obstacles, which has convergence guarantees. The placement of the reference point for such dynamic scenarios is ongoing research.}

\subsection{Future Work}
 Future work can extend the proposed work in the following areas:
\begin{itemize}
    \item \textbf{Low-level controller:} The low-level controller used in crowd navigation displaced the evaluation point away from the center of the robot. This resulted in an increased (conservative) margin around the robot. The way the shape of the robot is considered in the algorithm should be improved.
    \item \textbf{Environment recognition:} The update rate of the (deep-learning-based) tracker was approximately 5 Hz, while the present avoidance algorithm ran at a frequency of 50 Hz to 100 Hz. The obstacle avoidance was often evaluated with old environment information (but updated robot position). An intermediate estimator could predict how the crowds move in-between.
    \item \textbf{High-level planning:} The combination of the fast obstacle avoidance controller with slower planning algorithms could allow to handle more complex environments, i.e., including the avoidance of surrounding (non-star-shaped) environments.
    \item \textbf{Evaluation in high dimensional space}: The experimental implementation is executed on an autonomous wheelchair (a two-dimensional scenario). However, this work provides a theoretical solution, which can be applied to three and higher dimensional spaces. The next step will be the implementation and evaluation in a higher-dimensional space.
\end{itemize}

\section{Conclusion}
A dynamical system-based algorithm for local navigation under convergence constraint is presented. 
The work provides and tests the solution for local crowd navigation.
It ensures certain convergence constraints to not only safely navigate but also reach the goal in local scenarios. 
The advantage of the method comes from the low complexity and speed of the algorithm.
Furthermore, tuning free convergence is obtained in star-world scenarios.
This will allow to scale to higher dimensions and transfer to various scenarios. 

\section{ACKNOWLEDGEMENT}
We would like to thank the support of Diego Paez-Granados and David Gonon for the running of the experiments. Their effort and insight have contributed to the implementation using QOLO and the qualitative analysis of the results.

\appendix
\subsection{Directional Weighted Mean} \label{sec:dir_weighted_mean}
The weighted summation of vectors can result in a zero-sum (e.g., two vectors opposing each other with equal weight) and lead to undesired local-minima of dynamical systems. \\
We extend here the directional weighted summing introduced in \cite{huber2019avoidance} for more general application. It ensures that the summed vector field is free of local minima.
\begin{equation}
  \{\vect v \in \mathbb{R}^d \, :\,  \| \vect v\| = 1\} \label{eq:direction}
\end{equation}
The transformation into direction space $\mathcal{K}$ is given by:
\begin{equation}
\mathcal{K} = \{\kappa \in \mathbb{R}^{d-1} \; : \; \| \kappa \| < \pi \} \label{eq:dir_space}
\end{equation}
The direction space is with respect to a base vector $\vect{b}$ which is the first column of the orthonormal transformation matrix $\matr B$. This allows the transformation into the new basis:
\begin{equation}
  \hat{\vect v}_{i}=  \matr B ^T \vect{v}_{i} \label{eq:vector_trafo}
\end{equation}

The magnitude of the transformed vector in direction space is equal to the angle between the original vector and the reference vector. The transformation of the initial vector $\vect v_i$ in the direction-space is:
\begin{equation}
  \kappa_i(\vect{b}) = \vect k(\vect v_i, \vect b) =
  \begin{cases}
    \arccos \left(\hat{ \vect{v}}_{i [1]} \right)
    \frac{ \hat{\vect v}_{i [2:]} }{\|  \hat{\vect v}_{i [2:]} \|}  & \text{if} \;\; \hat{ \vect{v}}_{i [1]} \neq 1 \\
    \vect 0  & \text{if} \;\; \hat{ \vect{v}}_{i [1]} =1
\end{cases}
\label{eq:kappa_trafo}
\end{equation}

The mean is evaluated as a function of the weight $w_i$ of all $N^v$ vectors:
\begin{equation}
  \bar \kappa  = \sum_{i=1}^{N^v} w_i \kappa_i \label{eq:kappa_sum}
\end{equation}
The mapping into original space is evaluated as:
\begin{equation}
  \bar{\vect v}(\bar \kappa) =
  \begin{cases}
    \matr B \begin{bmatrix} 1 & 0 & .. & 0  \end{bmatrix}^T
    & \text{if} \;  \| \bar{\kappa} \| = 0  \\
    \matr B
    \begin{bmatrix}
      \cos \left(\| \bar{ \kappa}  \| \right) &
      \sin \left(\| \bar{\kappa} \|\right) \frac{\bar{\kappa}}{ \| \bar{\kappa} \|}  
    \end{bmatrix}^T
    &  \text{otherwise}
  \end{cases}
\label{eq:kappa_trafo_inv}
\end{equation}%

\subsubsection{Intuition}
\begin{figure}[t]\centering
\begin{subfigure}{.52\columnwidth} %
\centering
\includegraphics[width=\textwidth]{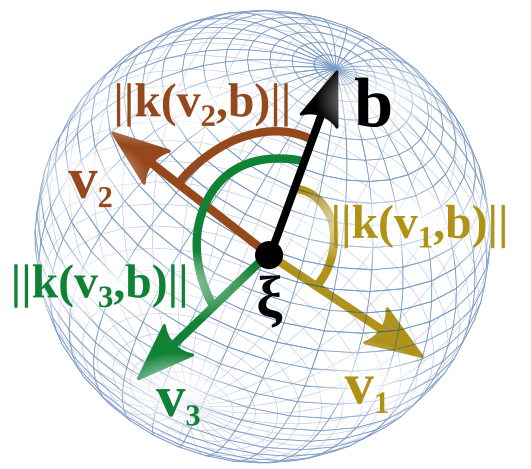}
\caption{Three initial vectors.}
\label{fig:modulatedDS}
\end{subfigure}%
\begin{subfigure}{.47\columnwidth} %
  \centering
  \vspace{2.3ex}
      \includegraphics[width=1.0\textwidth]{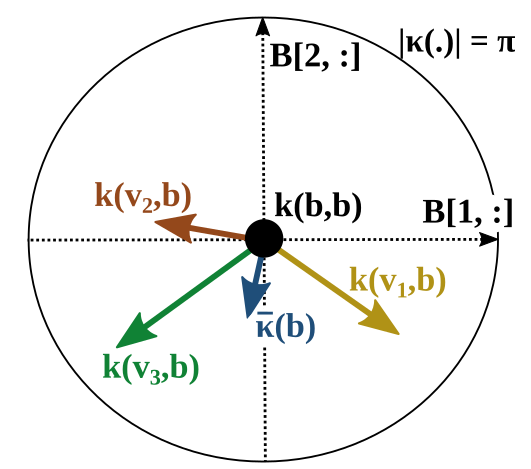}
\caption{Direction-space}
\label{fig:kappa-space}
\end{subfigure}
\caption{Various directions (here three) are described with respect to a basis direction $\vect r^0$ (a). The directions are transformed to the direction-space $\mathcal{K}$ (b) where the weighted mean, $\bar \kappa$, is obtained.}
\label{fig:interpolationDirection}
\end{figure}

In the two-dimensional case, this hyper-sphere is a line that represents the angle between the initial DS $\vect f(\xi)$ and the modulated DS $\dot \xi$. It has a magnitude strictly smaller than $\pi$, the directional space is a vector space, where the weighted mean is taken (see for the three dimensional case in Fig.~\ref{fig:interpolationDirection}). 
\\ \\
\noindent\textbf{Theorem A}
\textit{Consider a unit vector $\vect b$ as the basis for the projection given in (\ref{eq:kappa_trafo}) and the corresponding reconstruction function defined in (\ref{eq:kappa_trafo_inv}). The resulting transformation of unit vector $\vect k(\vect v, \vect b) \; : \;  \{ \vect v \in \mathbb{R}^d \setminus - \vect b \;\; : \;\; \|\vect{v}\| = 1 \} \rightarrow \mathcal{K}$ defined in (\ref{eq:dir_space}) is a bijection and the basis vector projects to the origin, i.e., $\vect b \rightarrow \vect 0$.} \\

\textbf{Proof: }
The proof is divided into three parts: \textit{(I)} showing that transformation and reconstruction are the inverse functions of each other, \textit{(II)} any unit vector is transformed to the direction space $\mathcal{K}$, and \textit{(III)} is reconstructed to a unit vector.

\textit{(I) Inverse Functions:} To be the inverse functions, applying one after the other onto a vector, must results in the original vector. (Vectors $\vect b $ will be treated separately below).   \\
Let us apply the forward transformation based on (\ref{eq:vector_trafo}) and (\ref{eq:kappa_trafo}) to a unit vector $\vect v_1$. It can be summarized to:
\begin{align}
  \kappa(\vect{b})  = \arccos\left(\langle \vect b, \vect v_1 \rangle \right) 
  \frac{ {\hat{\matr B}}^T \vect v_1}
  {\| \hat{\matr B} {\vect v}_1\|}
\end{align}
with $\hat{\matr B}$ the matrix $\matr B$ without the first row. \\
We apply it to a single vector, hence the corresponding weight is $w_1=1$. From (\ref{eq:kappa_sum}), we get $\bar \kappa = \kappa_1$. The reconstruction follows with (\ref{eq:kappa_trafo_inv}):
\begin{align*}
\bar {\vect v} & = \matr B
  \begin{bmatrix}
    \cos \left({\| \bar{ \kappa}  \|}\right) \\
    \sin \left( \| \bar{\kappa} \|\right) \frac{\bar{\kappa}}{ \| \bar{\kappa} \|}
  \end{bmatrix} 
  = 
  \matr{B}
  \begin{bmatrix}
    \cos \left(\arccos \left( \langle \vect b , \vect v_1 \rangle\right) \right) \\
  \frac{\sin(\arccos \left(\langle \vect b, \vect v_1 \rangle) \right) \left( \hat{\matr B}\right)^T {\vect v}_1}{\| \left( \hat{\matr B}\right)^T {\vect v}_1\|} 
  \end{bmatrix} \\
  & = 
  \matr{B} 
  \begin{bmatrix}
  \langle \vect b , \vect v_1 \rangle  \\
  {\left( \hat{\matr B}\right)^T {\vect v}_1}
  \end{bmatrix}
  = \matr{B}  \left( {\matr B}\right)^T {\vect v}_1
  = \vect v_1
\end{align*}
by using 
\begin{multline*}
  \sin(\arccos \left(\langle \vect b, \vect v_1 \rangle) \right)
  = \sqrt{1-\langle \vect b , \vect v_1 \rangle^2} \\
  = \sqrt{\| \matr B \vect v_1 \| ^2 - \langle \vect b , \vect v_1 \rangle^2}
  =  {\| \left( \hat{\matr B}\right)^T {\vect v}_1\|}
\end{multline*}
For the case that $\vect v_1 = \vect b$, we get from (\ref{eq:kappa_trafo}) that $\bar \kappa = \kappa_1 = \vect 0$. And with (\ref{eq:kappa_trafo_inv}, we get $\bar{\vect v}(\bar \kappa) = \vect b$, i.e., the original vector. \\
Hence for all cases, the transformation is bijective.\\
\\
\textit{(II) Transformation Domain} From definition of the transform in (\ref{eq:kappa_trafo}), the maximum magnitude is the $\arccos$, which is $\| \kappa_i\| < \pi$. Hence it lies in the domain of (\ref{eq:dir_space}). (The magnitude $\pi$ is only reached for a vector $-\vect b$, which is excluded from the transform).\\
\\
\textit{(III) Reconstruction Domain} From the inverse transform (\ref{eq:kappa_trafo_inv}), we have that norm of any transformed vector $\|\bar{\vect v}(\bar \kappa)\| = 1$, this follows from the fact that $\matr B$ is orthonormal. As a result they all lie in the domain (\ref{eq:direction}). 
${}$ \hfill $\blacksquare$ \\
\\
\subsubsection{Pairwise Closest Distance in Direction Space} \label{sec:gradient_descent}
\begin{figure}[t]\centering
\begin{subfigure}{.49\columnwidth} %
\centering
\includegraphics[width=\textwidth]{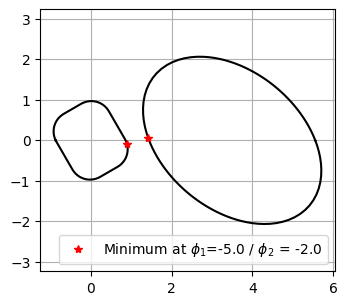}
\caption{Minimum Distance}
\label{fig:cloest_distance_ellipse_square}
\end{subfigure}%
\begin{subfigure}{.49\columnwidth} %
\centering
\includegraphics[width=\textwidth]{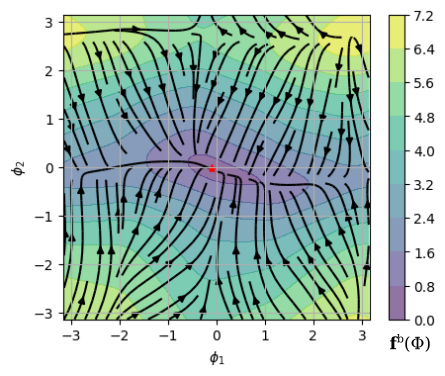}
\caption{Value Function}
\label{fig:gradient_descent_direction_space}
\end{subfigure}%
\caption{The minimum distance problem for a rectangular object (with margin) and an ellipsoid. The boundary-reference-point which corresponds to the closes point is marked in red (a) and the corresponding gradient descent problem in direction space (b).}
\label{fig:closets_distance}
\end{figure}

The directional space can be used for gradient descent to find the closest distance between two (convex) obstacles. This is done by moving along the surface of the obstacle in direction space. Since the direction space is of dimension $d-1$, finding the closest point of each obstacle has only $2(d-1)$ degrees of freedom (instead of $2d$ in the Cartesian space). The problem converges to the global minimum when the points start on the line which connects the two center points. \\
The optimization problem is given as:
\begin{equation*}
  \min_{\Phi} \, {\vect f\,}^b(\Phi)
  \qquad \text{with} \;\;
  {\vect f\,}^b(\Phi ) = \| \vect \xi^b_1(\phi_1) - \vect \xi^b_2(\phi_2) \|
  \, , \;\;
  \Phi =
  \begin{bmatrix} \phi_1 \\  \phi_2 \end{bmatrix} \label{eq:direction_space_value_function}
\end{equation*}
where the $\xi^b_i(\phi_i)$ denotes the boundary point in the direction $\phi_i$ for the obstacle $i$ with respect to its reference point $\xi^r_i$. The direction space of each obstacle is created such that the null-direction points towards the other obstacles center.
An example is visualized in Fig.~\ref{fig:closets_distance}.  \\

\subsubsection{Intersecting Obstacle Descent} \label{sec:intersecting_obstacle_descent}
\begin{figure} %
\centering
\includegraphics[width=0.99\columnwidth]{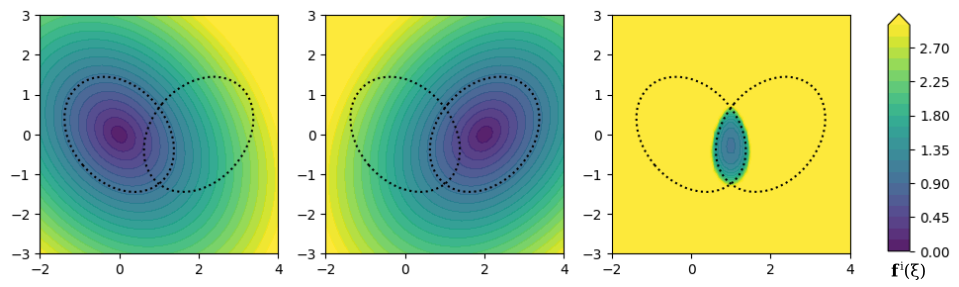}
\caption{The individual $\Gamma$-function from the obstacle in (a) and (b) are summed as described in (\ref{eq:intersection_summing}). The resulting value function (c) is used for the angle based gradient-descent.}
\label{fig:common_reference_point}
\end{figure}

Two intersecting obstacles need to share one reference point, which lies within both obstacles. The simplification of the problem to surface points only is not of use anymore. Hence, the optimization problem is evaluated in Cartesian space to find a common point that lies inside of the two obstacles' boundaries:
\begin{equation*}
  \min_{\xi \in \mathcal{X}_1^b \cap \mathcal{X}_2^b} \, \vect f^i(\vect \xi)
  \qquad \text{with} \;\;
  \vect f^i(\vect \xi) =  \frac{\Gamma^b}{\Gamma^b-\Gamma_1(\xi)} + \frac{\Gamma^b}{\Gamma^b-\Gamma_2(\xi)} \label{eq:intersection_summing}
\end{equation*}
with $\Gamma^b$ the base distance value, i.e. where the value function reaches infinity. It is chosen slightly larger than the $\Gamma$-value on the surface, we simply choose $\Gamma^b = 1.1$.
The step size can be optimized based on the gradient.
The optimization problem is convex, and points starting within the intersection region will stay inside due to the infinite repulsion at the boundary as $\Gamma \rightarrow 1$.
The value function of two ellipses can be found in Fig.~\ref{fig:common_reference_point}.\\

\subsubsection{Closest Distance for Mixed Environments}
The above method for gradient descent in directional space to find the closest distance between two objects, can be applied to an object-boundary pair, if obstacle's curvature $c^o$ is larger than boundary's curvature $c^b$  at any position:
\begin{equation}
  c^o (\xi_1) < c^b(\xi_2) \qquad \forall \, \xi_1, \, \xi_2 \label{eq:curvature_condition}
\end{equation}
with the local curvature being defined in (\ref{eq:curvature}). \\
Note that for non-circular obstacles, the condition might locally not hold mainly if the space contains polygon obstacles with local flat regions ($c=0$). This can lead to a locally non-optimal solution when placing the reference point based on the closest distance. \\

\subsection{Proof of Theorem 1} \label{sec:proofTheorem1}
\subsubsection{Applicability of General Proofs}
In \cite{huber2019avoidance} convergence has been proven for star-shaped obstacles. The proof was developed based on the distance function $\Gamma(\xi)$. Due to inverting the distance function for enclosing wall obstacles and a continuous definition of the modulation, the proof of star-shaped obstacles applies to the case of enclosing walls.
\subsubsection{Continuity Across Reference Point} \label{sec:continuousExtension}
In Sec.~\ref{sec:distance_inversion} the Inverted distance function $\Gamma^w(\xi)$ was not defined at the reference point, as it reaches an infinite value. The continuous definition for the eigenvalue is a unit value, i.e. $\lambda^e(\xi^r)=\lambda^r(\xi^r)=1$, it follows that the diagonal matrix is equal to the identity matrix $D(\xi^r) = I$. As shown in (\ref{eq:modulation_reference}23), we get:
\begin{equation}
  \xi=\xi^r \;\; \rightarrow \;\; \dot \xi = \matr E \, \matr D
  \, \matr E^{-1} \vect f(\xi^r)= \matr E \, \matr I \, \matr E^{-1} \vect f(\xi^r) = \vect f(\xi^r) \label{eq:continous_definition}
\end{equation}
i.e. no modulation of the initial DS. In fact, this is equivalent to the case far away for a classical obstacle with $\lim_{\|\xi-\xi^r\| \rightarrow \infty}\Gamma^o(\xi) \rightarrow \infty $. \\
Even though the basis matrix $\matr E(\xi)$ is not defined at $\xi^r$, the  DS is continuously defined across this point since the modulation has no effect. \\
The trajectory that traverses the reference point $\xi^r$ of the inverted obstacle corresponds to the trajectory that gets stuck in a saddle point for a common obstacle. As a result, there is full convergence for the inverted obstacles.

\subsection{Proof of Theorem 2}
We show first that the modulation has full rank and hence that the dynamics does not vanish outside the attractor and that it is smooth.
\label{sec:proofTheorem2}
\subsubsection{Full Rank} 
\begin{figure}[t]\centering
\centering
\includegraphics[width=\columnwidth]{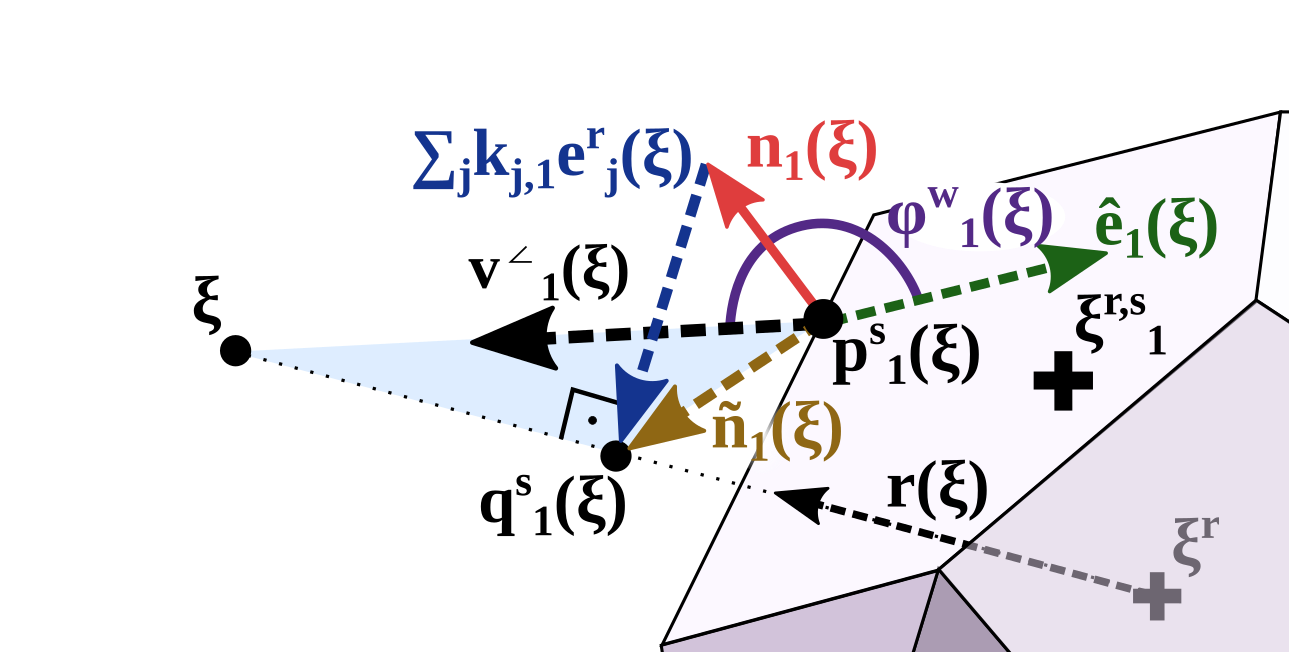}
\caption{Visualization of variables used for the weighted directional mean.}
\label{fig:inequality_triangle}
\end{figure}

The basis matrix from (\ref{eq:basis_matrix_nonsmooth}) has full rank everywhere outside of the obstacle, if the following condition holds:
\begin{equation}
   \arccos \left( \langle \vect r(\xi), \, \hat{\vect n}(\xi)  \rangle \right) < \pi/2 \label{eq:full_rank_condition}
\end{equation}
The angle between the normal to each surface i, $\vect n_i(\xi)$ and the reference direction $\vect r (\xi)$ can be evaluated by defining an vector $\tilde{\vect n}_i(\xi) = \vect n_i(\xi) + \sum_{j=1}^{d-1} k^e_i\vect e_j(\xi)$ with $\langle \vect e_j(\xi), \, \vect r(\xi) \rangle = 0$, $k^e_i \in \mathbb{R}$ such that $\vect p_i^s(\xi) + \tilde{\vect n}_i(\xi)$ intersects with $\xi^r + k^r \vect r(\xi)$ at $\vect q^s_i(\xi)$ with $k^r \in \mathbb{R}$ (Fig.~\ref{fig:inequality_triangle}). \\
This allows to create a triangle spanned by the lines  $\xi$, $\vect p_i^s(\xi)$ and $\vect q^s_i(\xi)$, colored in blue in Fig.~\ref{fig:inequality_triangle}. Using the associative law of the dot product, the geometry constraint of the blue triangle and (\ref{eq:angle_weight}), the maximum angle results in:
\begin{equation}
 \langle   \vect{n}_i,  \vect r \rangle = \langle \tilde{\vect{n}}_i, \vect r \rangle \geq  \langle \xi-\vect p_i^s, \vect r \rangle \geq 0     \quad \forall \, w_i(\xi) > 0
\end{equation}
\noindent
Hence the directional transformation of (\ref{eq:kappa_trafo}) results in $\| \kappa_i \| < \pi/2\|, \; \forall \, w_i(\xi) > 0$.
Using additionally the \textit{triangle equality} for vectors:
 $ \| \kappa _1 +  \kappa_2 \| \leq \| \kappa_1 \| + \| \kappa_2 \| $
applied to all surface directions, it follows with (\ref{eq:kappa_sum}) that:
\begin{align}
  \| \bar \kappa \| & = \| \sum_{i=1}^{N^v} w_i \kappa_i \| \leq \sum_{i=1}^{N^v} w_i \| \kappa_i \| \leq \left( \sum_{i=1}^{N^v} w_i \right)\| \max_{i \; \text{with} \; w_i > 0}\kappa_i \|  \leq \frac{\pi}{2} \nonumber
\end{align}
Since the basis vector of the directional mean is $\vect r(\xi)$, with (\ref{eq:kappa_trafo_inv}) condition (\ref{eq:full_rank_condition}) holds true.

\subsubsection{Smooth Vector Field}
 The continuous extension across the reference point is defined in Appendix~\ref{sec:continuousExtension} and applicable, too. \\
The reference direction $\vect r(\xi)$ and the distance function $\Gamma(\xi)$ does not have any other discontinuity. \\
The pseudo normal $\hat{\vect n}(\xi)$ is smoothly defined across space. Even in the case when the edge point with the minimum is switching (\ref{eq:shortest_distance}), no discontinuity occurs since the angle will stay the same due to the flat surface.

\subsubsection{Applicability of General Proofs}
Since we have a smooth field of normal vectors $\vect n(\xi)$, we further need to define any smooth distance function which decreases its value with increasing distance. The two properties are sufficient to comply with the proof of Sec.~\ref{sec:proofTheorem1}
${}$ \hfill $\blacksquare$

\subsection{Proof Theorem 3} \label{sec:proofTheorem3}
As an agent approaches the surface of an obstacle, the importance weight from this obstacle is approaching one, i.e., $\lim_{\Gamma_{\hat{o}}(\xi) \rightarrow 1} w_{\hat{o}}=1$, see (\ref{eq:obtacle_weight}). It follows with (\ref{eq:relative_obstacle_velocity}) that the relative velocity is $\dot{\tilde \xi} = \dot{\tilde \xi}^{\mathrm{tot}} =  \dot{\tilde \xi}_{\hat o}$. \\
As a result it is sufficient to analyze impenetrability for each obstacle individually. The next step is to ensure impenetrability of the three cases in (\ref{eq:safe_velocity}):

\subsubsection{Evaluation in Moving Frame: } $\dot{\xi}  = \dot{\xi} $\\
The simplest case comes with no stretching, but the evaluation in the local frame of the moving boundary of the obstacle. It follows that the Neuman-boundary condition for impenetrability holds (see also \cite{huber2019avoidance}). 

\subsubsection{Contraction within Margin} $\dot{\xi}  = v^{\mathrm{max}} \| \dot{\xi} \| \dot{\xi}$ \\
This contraction is only performed, if it results in a normal velocity which is larger than the velocity of the obstacle $\dot{\tilde{\xi}}$, i.e. the evaluation in the moving frame results $\langle (\dot{\xi}-\dot{\tilde{\xi}}, \vect{n}(\xi) \rangle \geq 0$, hence ensuring impenetrability.

\subsubsection{Contraction in Tangent Direction} %
\begin{equation*}
  \dot{\xi} = v^n \vect{n}(\xi) + \sqrt{\left(v^{\mathrm{max}}\right)^2 -\|\dot{\xi}^n\|^2} \; \vect{e} (\xi)
\end{equation*} %
This limited contraction along the normal direction ensures that the velocity in normal direction remains equal to the obstacles' velocity. The evaluation of the Neuman boundary condition in the moving frame leads to:
\begin{equation*}
\langle (v^n \vect{n}(\xi) + \sqrt{\left(v^{\mathrm{max}}\right)^2 -\|\dot{\xi}^n\|^2} \; \vect{e} (\xi) )- \tilde{\xi}, \vect{n}(\xi) \rangle
=  v^n - v^n = 0
\end{equation*}
using the definition of (\ref{eq:max_velocity}) and the fact that the normal $\vect{n}(\xi)$ and the tangent $\vect{e}(\xi)$ are orthogonal.\\
\\
Hence, we have impenetrability for multiple obstacles. \\
${}$ \hfill $\blacksquare$

\subsection{Proof Theorem 3} \label{sec:proofTheorem3}
As an agent approaches the surface of an obstacle, the importance weight from this obstacle is approaching one, i.e., $\lim_{\Gamma_{\hat{o}}(\xi) \rightarrow 1} w_{\hat{o}}=1$, see (\ref{eq:obtacle_weight}). It follows with (\ref{eq:relative_obstacle_velocity}) that the relative velocity is $\dot{\tilde \xi} = \dot{\tilde \xi}^{\mathrm{tot}} =  \dot{\tilde \xi}_{\hat o}$. \\
As a result it is sufficient to analyze impenetrability for each obstacle individually. The next step is to ensure impenetrability of the three cases in (\ref{eq:safe_velocity}):

\subsubsection{Evaluation in Moving Frame: } $\dot{\xi}  = \dot{\xi} $\\
The simplest case comes without stretching, but the evaluation in the local frame of the moving boundary of the obstacle. It follows that the Neuman-boundary condition for impenetrability holds (see also \cite{huber2019avoidance}). 

\subsubsection{Contraction within Margin} $\dot{\xi}  = v^{\mathrm{max}} \| \dot{\xi} \| \dot{\xi}$ \\
This contraction is only performed, if it results in a normal velocity which is larger than the velocity of the obstacle $\dot{\tilde{\xi}}$, i.e. the evaluation in the moving frame results $\langle (\dot{\xi}-\dot{\tilde{\xi}}, \vect{n}(\xi) \rangle \geq 0$, hence ensuring impenetrability.

\subsubsection{Contraction in Tangent Direction} %
\begin{equation*}
  \dot{\xi} = v^n \vect{n}(\xi) + \sqrt{\left(v^{\mathrm{max}}\right)^2 -\|\dot{\xi}^n\|^2} \; \vect{e} (\xi)
\end{equation*} %
This limited contraction along the normal direction ensures that the velocity in normal direction remains equal to the obstacles' velocity. The evaluation of the Neuman boundary condition in the moving frame leads to:
\begin{equation*}
\langle (v^n \vect{n}(\xi) + \sqrt{\left(v^{\mathrm{max}}\right)^2 -\|\dot{\xi}^n\|^2} \; \vect{e} (\xi) )- \tilde{\xi}, \vect{n}(\xi) \rangle
=  v^n - v^n = 0
\end{equation*}
using the definition of (\ref{eq:max_velocity}) and the fact that the normal $\vect{n}(\xi)$ and the tangent $\vect{e}(\xi)$ are orthogonal.\\
\\
Hence, we have impenetrability for multiple obstacles. \\
${}$ \hfill $\blacksquare$

\editcolor{\subsection{Evaluation Weights for a Robot Arm}\label{sec:section_link_weight}
We introduce link weights $w_l^L$ for each link $l \in [1 .. N^L]$, and section weights $w_{l,s}^S$ for each section point $\xi_{l,s}^S$ with $s \in [1, .. N^S]$. The weights determine how much the obstacle avoidance command influences the control, i.e., zero weights indicate following the goal velocity only, whereas a weight of 1 means full avoidance at this specific position. \\
These weights are designed, such that their product is smaller than one, i.e.:
\begin{equation*}
  0 \leq \sum_{l} \left( w^L_l \sum_{s} w^S_{l,s} \right) \leq 1 \label{eq:weight_sum_1} 
\end{equation*}
Furthermore, if a section point approaches the surface of an obstacle, it should dominate, i.e.:
\begin{equation*}
  \Gamma^d(\xi_{l, s}^S) \rightarrow 1
  \quad \Rightarrow \quad
  w_l^L w_{l, s}^S \rightarrow 1
\end{equation*}
And when the robot arm is far away from any surface, all weights should go to zero:
\begin{equation*}
  \min_{l, s}\left(\Gamma(\xi_{l,s}^S)\right) \rightarrow \infty
  \quad \Rightarrow \quad
  \sum_{l} \left( w^L_l \sum_{s} w^S_{l,s} \right) \rightarrow 0
\end{equation*}

\subsubsection{Base Weight} \label{sec:link_weights}
For each link $l$ and section point $s$, we define a $\Gamma$-danger weight $w^{\Gamma}(\Gamma^d):  \left] 1, \infty \right[ \rightarrow \left] \infty, 0 \right[$ to represent the danger of colliding for a specific point: the higher the weight, the greater the chance of collision. It is defined as:
\begin{equation}
    w^{\Gamma}(\Gamma^d) =
    \begin{cases}
    \frac{\Gamma^c - \Gamma^{\mathrm{min}}}{\Gamma^d(\xi_{s, l}^S) - \Gamma^{\mathrm{min}}} - 1 & \text{if}  \;\; \quad \Gamma^d(\xi_{s, l}^S)  < \Gamma^c\\
    0 & \text{otherwise}
    \end{cases}
    \label{eq:gamma_weights}
\end{equation}
where $\Gamma^{\mathrm{min}} = 1$  is the lower bound, $\Gamma^c > 1$ the cutoff value and $\Gamma^d(\xi)$ defined in \ref{eq:danger_gamma}.\\

\subsubsection{Link Weights} \label{sec:link_weights}
The influence of each link $l$ is evaluated as:
\begin{equation}
    \hat{w}^L_l = c^L |\dot{\vect{q}}^g_{[l]} | \frac{l}{N^L} \max_{s \in [1 .. N^S]} w^{\Gamma}(\xi_{s, l}^S)
    \quad \forall \, l \in [1 .. N^L] \label{eq:joint_weight}
\end{equation}
where $c^L \in \mathbb{R}_{>0}$ is a constant link-weight factor and $\dot{\vect{q}}^g_{[l]}$ the $l$-th element of the goal vector defined in (\ref{eq:goal_command}). The $l$-factor in the equation gives an increasing importance for links closer to the end-effector. \\
The link weights are obtained through normalization:
\begin{equation}
  w^L_l =
  \begin{cases}
    \hat{w}^L_l / \hat w^{\mathrm{sum}} \quad & \text{if} \; \hat w^{\mathrm{sum}} > 1 \\
    \hat{w}^L_l & \text{otherwise}
  \end{cases}
  \;\;\; \text{with} \;\;
  \hat w^{\mathrm{sum}} = \sum_{l=1}^{N^L} \hat w^L_l
  \label{eq:link_weight}
\end{equation}

\subsubsection{Section Weights} \label{sec:section_weights}
For all links $l$ with $w_l^L > 0$, the influence weight of each section point is evaluated as:
\begin{equation}
  w^S_{s} = \frac{\hat{w}^S_{s}}{\sum_{s=1}^{N^S} \hat w^S_{s}}
  \quad \text{with} \quad
  \hat{w}^S_{s} = \frac{s}{N_S} w^{\Gamma}(\xi_{l, s}^S)
    \quad \forall \; s \in [1 .. N_S] \label{eq:point_weight}
\end{equation}
}
\renewcommand*{\bibfont}{\small}
\printbibliography


\end{document}